\documentclass[journal]{IEEEtran}

\usepackage{amsmath}
\usepackage{amsfonts}
\usepackage{amssymb}

\usepackage{amsthm}
\usepackage{mathtools}
\usepackage{tabularx,booktabs}
\usepackage{graphicx}
\usepackage{subfigure}
\usepackage{enumerate}
\usepackage{float}
\usepackage{url}
\usepackage{verbatim}
\usepackage{cite}
\usepackage{diagbox}
\usepackage{multirow}
\usepackage{makecell}
\usepackage{algorithm}  
\usepackage{algorithmicx}  
\usepackage{algpseudocode} 
\usepackage{bbding}
\usepackage{bm}
\usepackage{cases}
\usepackage[linkcolor=black,citecolor=black,urlcolor=black,colorlinks=true]{hyperref}
\usepackage{tikz}
\usepackage{etoolbox}
\usepackage{graphicx}

\usepackage{pifont}
\usepackage{threeparttable}

\newcommand{\circled}[2][]{\tikz[baseline=(char.base)]
	{\node[shape = circle, draw, inner sep = 1pt]
		(char) {\phantom{\ifblank{#1}{#2}{#1}}};%
		\node at (char.center) {\makebox[0pt][c]{#2}};}}
\robustify{\circled}

\graphicspath{{figures/}}
\IEEEoverridecommandlockouts

\newcommand{\df}[1]{\mathrm{d}{#1}}

\newcommand{\norm}[1]{\Vert{#1}\Vert}

\author{ Yuman Gao$^{*1}$, Jialin Ji$^{*1}$, Qianhao Wang$^{1}$, Rui Jin$^{1}$, Yi Lin$^{2}$, Zhimeng Shang$^{2}$, \\Yanjun Cao$^{1}$, Shaojie Shen$^{3}$, Chao Xu$^{1}$,  and Fei Gao$^{\dagger1}$
    \thanks{ $^*$Indicates equal contribution. }
	\thanks{$^\dagger$Corresponding author: {\tt\small fgaoaa@zju.edu.cn}.}
	\thanks{
	$^1$Key Laboratory of Industrial Control Technology, Institute of Cyber-Systems and Control, Zhejiang University, Hangzhou 310027, China, and the Huzhou Institute, Zhejiang University, Huzhou 313000, China.
	}
    \thanks{
    $^2$DJI Technology Co., Shenzhen 510810, China.
	}
    \thanks{
    $^3$Department of Electronic and Computer Engineering, Hong Kong University of Science and Technology, Hong Kong, China.
	}
    \thanks{
	This work was supported by the National Natural Science Foundation of China under grant no. 62322314, the DJI-ZJU FAST Autonomous Drone Research Funding, and the Fundamental Research Funds for the Central Universities.
	}

}

\title{\LARGE \bf Adaptive Tracking and Perching for Quadrotor\\ in Dynamic Scenarios}

\begin{document}
    \maketitle

\begin{abstract}
Perching on the moving platforms is a promising solution to enhance the endurance and operational range of quadrotors, which could benefit the efficiency of a variety of air-ground cooperative tasks. 
To ensure robust perching, tracking with a steady relative state and reliable perception is a prerequisite. 
This paper presents an adaptive dynamic tracking and perching scheme for autonomous quadrotors to achieve tight integration with moving platforms.
For reliable perception of dynamic targets, we introduce elastic visibility-aware planning to actively avoid occlusion and target loss.
Additionally, we propose a flexible terminal adjustment method that adapts the changes in flight duration and the coupled terminal states, ensuring full-state synchronization with the time-varying perching surface at various angles.
A relaxation strategy is developed by optimizing the tangential relative speed to address the dynamics and safety violations brought by hard boundary conditions.
Moreover, we take SE(3) motion planning into account to ensure no collision between the quadrotor and the platform until the contact moment.
Furthermore, we propose an efficient spatiotemporal trajectory optimization framework considering full state dynamics for tracking and perching.
The proposed method is extensively tested through benchmark comparisons and ablation studies. To facilitate the application of academic research to industry and to validate the efficiency of our scheme under strictly limited computational resources, we deploy our system on a commercial drone (DJI-MAVIC3) with a full-size sport-utility vehicle (SUV). We conduct extensive real-world experiments, where the drone successfully tracks and perches at 30~km/h (8.3~m/s)  on the top of the SUV, and at 3.5~m/s with 60° inclined into the trunk of the SUV.

\end{abstract}


\section{Introduction}
\label{sec:intro}

\IEEEPARstart{U}{ncrewed} aerial vehicles (UAVs) that operate in conjunction with mobile platforms have recently emerged in a variety of applications, such as truck-drone delivery systems \cite{das2020synchronized,wu2022collaborative,liu2020two}, car-drone inspection systems \cite{wu2015coordinated,tokekar2016sensor,shen2017collaborative}, and air-ground search and rescue systems \cite{delmerico2017active,wu2022distributed,mittal2022vision}.
In these applications, due to the short flight duration of UAVs, they need to frequently perch on the mobile platform for resting or recharging, which necessitates the platform to come to a complete stop or interrupt its ongoing operations.
For higher automation and task efficiency, it is necessary to let the UAV smoothly attach to moving platforms without interruption.
This demands the UAV to track the moving and various angled platforms with steady relative state and reliable perception, and perch on it at a proper moment.
For an ideal planner promising successful tracking and perching applied in dynamic real-world scenarios, \textit{adaptability} must be explicitly considered and modeled to cope with several practical challenges analyzed as follows.

\begin{figure}[t]
    \begin{center}
        \includegraphics[width=1.0\columnwidth]{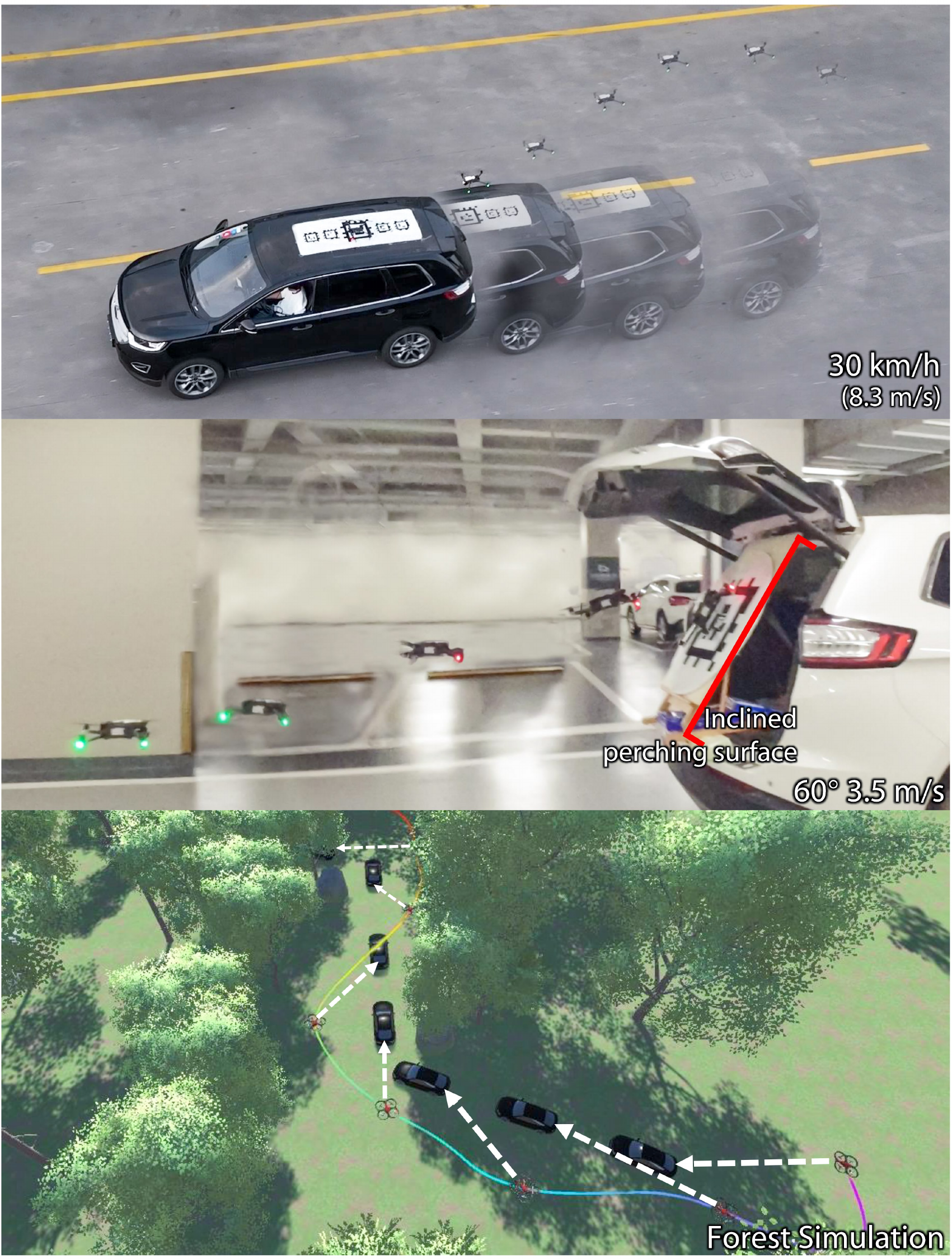}
    \end{center}
    \vspace{-0.3 cm}
    \caption{Simulations and real-world experiments of our adaptive tracking and perching system. Please watch our attached videos for more information at: youtu.be/5XKm7qkp2Xs, youtu.be/fBwW93Zq9ss.}
    \vspace{-0.5 cm}
    \label{fig:head}
\end{figure}

The first challenge is how to maintain stable observations during agile flight.
For a quadrotor chasing a high-speed ground vehicle, the non-cooperative movements of these two agents result in fluctuating relative motions. 
Such relative motion easily makes the target out of the limited field of view (FoV) of the drone's sensory device, causing target loss.
Moreover, irregular obstacle distributions or dynamic objects often block direct observations from the quadrotor to its target, producing occlusion.
To avoid the above situations, the trajectory of the drone should be adjusted adaptively according to the target movement and the surrounding environments, summarized as \textit{visibility} requirement.

The second challenge is how to smoothly and quickly attach to the dynamic platform.
A desired perching requires the quadrotor to synchronize its position, velocity, and attitude adaptively with the perching surface at a proper contact moment.
Since a quadrotor's attitude and motion are coupled, a time-varying terminal state introduces cross-dependent temporal and spatial conditions for the quadrotor's trajectory. 
To cope with such conditions, the quadrotor should flexibly adjust its flight duration and the coupled terminal states during trajectory generation.
Furthermore, strict terminal constraints may contain conflicts with the quadrotor's safety and dynamics, resulting in no feasible solution. 
Therefore, a relaxation approach is hoped for reasonably adjusting the final perching state. 
The above requirements are summarized as \textit{flexibility}.

The third challenge is how to react to sudden changes in complicated situations.
In a dynamic scenario without global information, obstacles that appear suddenly, target states that change quickly, and external disturbances that act severely would make the latest generated trajectory deprecated.
To let the quadrotor react adaptively in time, high-frequency replanning based on the onboard perception is necessary.
However, for a quadrotor with limited onboard resources, it's difficult to satisfy the planning efficiency, the modeling fidelity, and the solving completeness at the same time.
Therefore, the \textit{responsiveness} is another critical requirement for planning.

Apart from the above requirements, \textit{safety} and \textit{dynamic feasibility} are also fundamental considerations for conducting ideal aerial tracking and perching.
The generated trajectory should avoid any possible collisions with obstacles, and can only contact the target till the final moment. 
Moreover, perching maneuvers with large attitudes often push the actuators toward their physical limits, making dynamic feasibility essential. However, constraints on high-order dynamic states usually demand high computation. 

Based on the above observations and analysis, and built upon our previous conference papers \cite{ji2022real,wang2021visibility}, we propose a complete aerial system with an adaptive tracking-perching scheme for dynamic targets. 
In our proposed system, to actively enhance \textit{visibility}, we design a series of differentiable metrics considering the occlusion of obstacles, relative distance and angle, with the limited 3-D FoV of quadrotors. Based on the surrounding environment, the tracking distance can be elastically adjusted.
Moreover, to maintain high-quality observation, the quadrotor's position and attitude are jointly adjusted to lock the target centrally in the image space while perching.
Secondly, to provide \textit{flexibility} of determining the flight duration and the coupled terminal states, we propose a flexible terminal adjustment approach to ensure full-state synchronization with the time-varying perching surface.
This approach eliminates the terminal constraints along with reducing optimization variables.
Additionally, a relaxation strategy is developed by optimizing the tangential relative speed, addressing the conflict between terminal restrictions, safety, and dynamic feasibility.
Thirdly, to ensure \textit{safety} around the contact moment precisely, we geometrically model the quadrotor and platform to prevent collision, and further construct a concise geometric constraint.
Furthermore, \textit{dynamic feasibility} is guaranteed by constrained on high-order states, including angular velocity and thrust, with an efficient flatness mapping.
To collectively involve the above aspects in trajectory generation, we propose an efficient spatiotemporal trajectory optimization framework. Concise metrics and compact trajectory representation benefit the solving efficiency, ensuring \textit{responsiveness}.
Finally, to facilitate the application of academic research to industry and to validate our system in the real world, we deploy our adaptive tracking and perching scheme on a commercial drone (DJI-MAVIC 3) with a sport-utility vehicle (SUV) as the moving platform. 
Efficient trajectory optimization enables high-frequency replanning even in embedded processors with severely limited resources.
We present experiments in a variety of dynamic scenarios partly shown in Fig.~\ref{fig:head}, including successful tracking and perching on the SUV with a speed up to $30 \, \text{km/h}$ ($8.3 \, \text{m/s}$). 
Contributions of this paper are listed as:
\begin{enumerate}
    \item A series of differentiable planning metrics that enable visibility awareness against occlusion and target loss in aggressive flight efficiently.

    \item A concise approach that flexibly adjusts the terminal perching states with safety and feasibility guaranteed by terminal constraint relaxation.

    \item An efficient trajectory optimization framework considering full state dynamics and complex collision constraints for tracking and perching.
    
    \item A variety of simulations and real-world tests that validate the proposed methods with a commercialized quadrotor.  
\end{enumerate}

This paper consolidates the preliminary conference papers presented in~\cite{ji2022real, wang2021visibility} with significant functionality extension, performance improvements, as well as application promotion.
Compared to our previous work~\cite{ji2022real} which relies on external facilities for perception, we build a fully autonomous tracking and perching system with solely onboard sensors.
To better exploit the limited sensing capability of onboard vision, we propose a visibility-aware planner to deform the trajectory for improving the target's observation quality.
Besides, our previous tracking method~\cite{wang2021visibility} employs a decoupled way to generate a trajectory, where the temporal profile is optimized after its spatial shape.
In this paper, we jointly optimize the trajectory in space and time and result in much higher optimality. 
Finally, by deploying our method in a commercial quadrotor and testing it with a full-sized vehicle, we demonstrate that our system can be successfully deployed outside the laboratory. 
We consider this as an important step forward from our previous works which are only validated in human-controlled environments.

\section{Related Works}
\label{sec: related works}

\subsection{Aerial Tracking System}
\label{subsec: visibility-aware trajectory planning}
Several previous vision-based tracking works \cite{cheng2017autonomous,kendall2014board} formulate the trajectory planning and control integrally as a reactive control problem, and take the tracking error defined on image space as the feedback. These reactive methods can achieve real-time performance but are too short-sighted to consider safety and occlusion constraints. To overcome these drawbacks, model predictive control (MPC) planners \cite{nageli2017real,penin2018vision} and differential-flatness-based planners \cite{thomas2017autonomous, bonatti2020autonomous,han2021fast,jeon2019online,jeon2020integrated,ji2022elastic} with receding horizon formulations emerge.
Nageli \cite{nageli2017real} designs a modular visibility cost function based on the re-projection error of targets, and integrates it into an MPC planner. However, this method has a strong assumption that the obstacles are elliptical and thus cannot avoid arbitrarily shaped obstacles well.
Penin \cite{penin2018vision} formulates a non-linear MPC optimization problem with a visibility term in image space. Lacking environmental perception, this approach requires complete knowledge of the environment. 
Optimizing trajectories in flat-output space, Bonatti \cite{bonatti2020autonomous} trades off shot smoothness, safety, occlusion, and cinematography guidelines. Nevertheless, they avoid occlusions by making the connection line between the target and robot obstacle-free, neglecting the conical FoV shape of sensors.
Han \cite{han2021fast} proposes a safe tracking trajectory planner consisting of a kinodynamic searching front-end and a spatiotemporal optimal trajectory planning back-end. Without visibility considerations, it is prone to fail due to the target being out of the FoV or too close in relative distance.
Jeon \cite{jeon2019online, jeon2020integrated} proposes a graph-search-based preplanning procedure to cover safety and visibility, but carries out path smoothing afterward. Such inconsistency makes the trajectories may not meet all the constraints.
Ji \cite{ji2022elastic} uses a series of 2-D fans to represent the visible region for tracking planning. But the fans are only generated at the specified height, making this method pseudo-3D. 
These recent works address that visibility is significant in determining tracking success rate and robustness.
In contrast with the above works, our proposed differentiable metrics comprehensively consider the factors that affect visibility, including the 3-D shape of the conical FoV, the relative observation distance and angle, and obstacle occlusion.

\subsection{Dynamic Landing and Perching}
\label{subsec: Dynamic Landing}
For clarity, we refer to the methods that can merely adjust the end position and velocity as `landing`, and methods that can additionally adjust the end attitude as `perching`.
Dynamic landing on a moving platform has been widely studied in UAV research. 
Previous studies typically adopt control-level methods, which directly construct a state deviation term based on target observation. Controllers designed with the proportional–integral–derivative method or Lyapunov-function-based methods are widely adopted \cite{borowczyk2017autonomous,qi2019autonomous,ghommam2017autonomous}.
Besides, image-based visual servoing (IBVS) method \cite{serra2016landing,thomas2014toward,lee2012autonomous} is another group of control-level methods that define the visual error in image space with less computational complexity. 
However, the above methods are often shortsighted and face difficulties to introduce state constraints related to interacting objects. 
To compensate for these drawbacks, most of the recent works address dynamic landing issues with MPC \cite{maces2017autonomous,mohammadi2020vision,paris2020dynamic} or differential-flatness-based trajectory planning approaches \cite{falanga2017vision,wang2022quadrotor}, both with receding planning horizon.
However, all these above studies only aim to reach the landing position, lacking the ability to adjust desired attitudes at the contact moment.
Furthermore, the coordinated landing of quadrotors with moving platforms is also widely studied \cite{daly2015coordinated,muskardin2016landing}, where the joint planning and control are designed to accomplish rendezvous and landing. 
These methods require control of the platform introducing higher system complexity, and have limited application scenarios.

Many previous perching works \cite{tsukagoshi2015aerial, daler2013perching,kalantari2015autonomous,barrett2018autonomous, thomas2016aggressive,mao2023robust,paneque2022perception} study the problem of perching on stationary inclined surfaces. Most of these works \cite{tsukagoshi2015aerial, daler2013perching,kalantari2015autonomous,barrett2018autonomous} rely on particular mechanisms, such as suction or adhesive grippers, to bypass the terminal attitude requirement. Thomas \cite{thomas2016aggressive} proposes a planning and control strategy fully considering actuator constraints and formulates a quadratic programming (QP) problem using a series of linear approximations. However, this approach cannot adjust the perching duration, and the linearization is oversimplified. Based on this approach, Mao \cite{mao2023robust} proposes a global-bound-checking method to check the dynamic feasibility efficiently, then increases the trajectory duration and recursively solves a QP problem until actuator constraints are satisfied. However, the method ignores that the spatial profile of trajectories also affects dynamic feasibility, which can result in this constraint being unsatisfied.
Moreover, Paneque \cite{paneque2022perception} formulates a discrete-time multiple-shooting nonlinear programming (NLP) problem for perching on powerlines, with a task-specific perception-aware term. Nevertheless, this method adjusts terminal states by computationally-expensive multiple shooting, resulting in an excessive computation time for orders of magnitude longer than ours.

For dynamic perching, Vlantis\cite{vlantis2015quadrotor} studies the problem of landing a quadrotor on an inclined moving platform. By solving a discrete time non-linear MPC, the drone approaches the platform, while maintaining it within the camera’s FoV, and finally perches on it. However, such a computationally demanding problem is computed on a ground station. 
Moreover, the quite small inclined angle ($27^\circ$) and quite slow speed ($0.5 \, \text{m/s}$) are inadequate for validation. Hu \cite{hu2019time} achieves the differentially-flat spatiotemporal optimization considering the actuator constraints and collision constraints. But it is only for 3-DOF quadrotors in the 2-D coordinate system.
Liu \cite{liu2023hitchhiker} designs a suction cup gripper for quadrotors to perch on a moving target, but this work focuses more on the novel mechanical design and simply uses minimum-jerk-based trajectory generation without safety, actuator limit, and visibility considerations.


\begin{figure*}[t]
    \begin{center}
        \includegraphics[width=2.0\columnwidth]{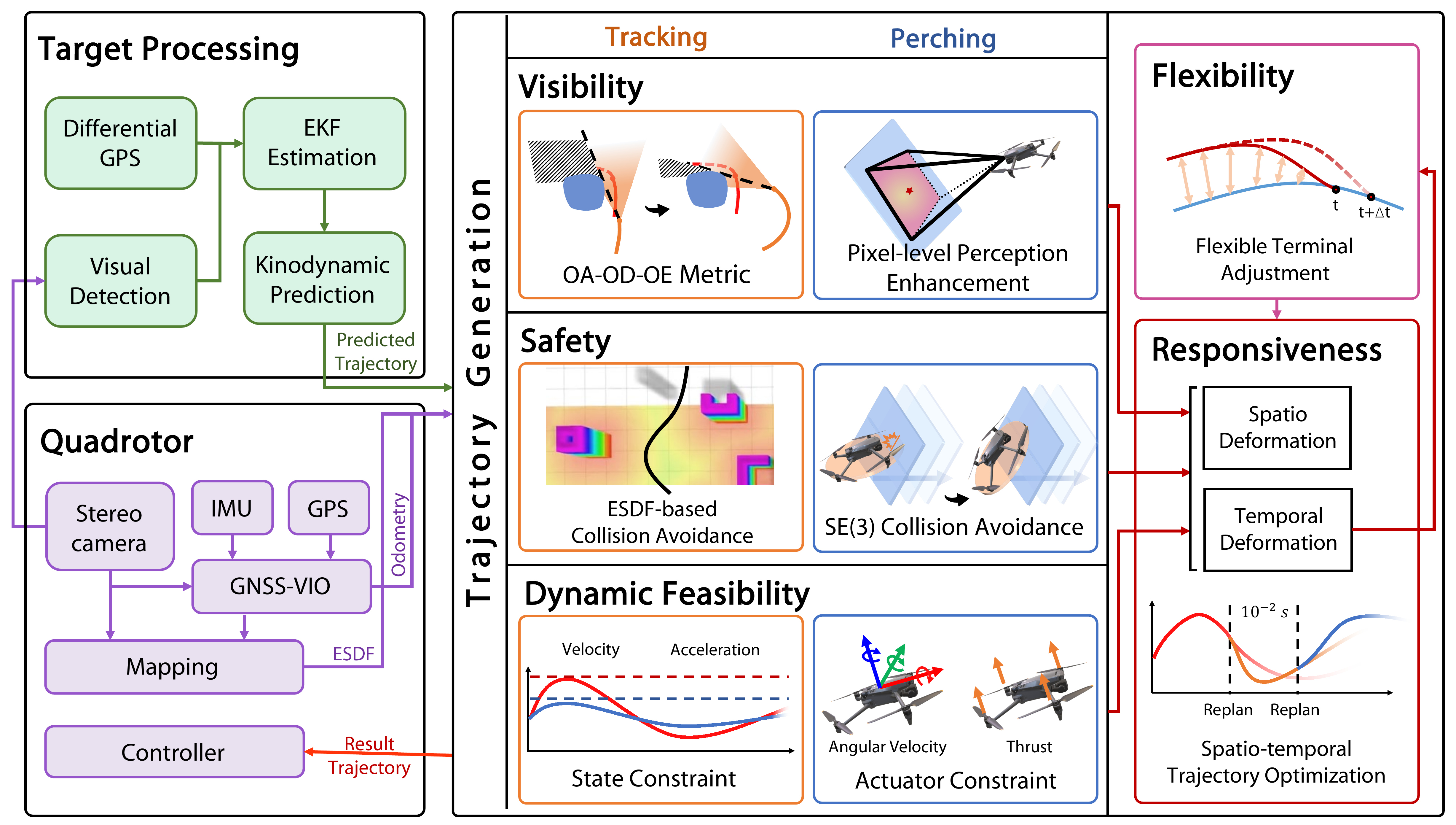}
    \end{center}
    \vspace{-0.3 cm}
    \caption{An overview of our complete aerial system with dynamic tracking and perching scheme. The trajectory generation module takes all the requirements mentioned in Sec. \ref{sec:intro} into account and provides spatiotemporal optimal feasible trajectory for stable tracking and dynamic perching.}
    \vspace{-0.2 cm}
    \label{fig:system}
\end{figure*}

\section{System Overview and Preliminaries}
\label{sec:system}
\subsection{System Architecture}
The overall architecture of our aerial system with the adaptive tracking and perching scheme is illustrated in Fig.~\ref{fig:system}. The target position is obtained by the fusion of coarser
differential-GPS-based relative localization (meter-level
error) and finer visual detection (centimeter-level
error).
After obtaining the target position, we conduct an extended Kalman filter (EKF) based estimation and predict the target trajectory in the future (Sec. \ref{subsec:Estimation model}). 
To meet the five aspects of requirements stated in Sec.\ref{sec:intro}, we design comprehensive constraints of visibility, safety and dynamic feasibility, respectively for tracking (Sec. \ref{sec:Visibility-aware Tracking Metrics}) and perching (Sec. \ref{sec:Robust and Dynamic Perching Metrics}). 
Collectively considering these constraints, we simultaneously optimize the spatial and temporal profile of the trajectory with high efficiency (Sec. \ref{sec:trajectory optimization}). 
When trajectory duration changes in temporal deformation, a flexible terminal adjustment approach is adopted to adaptively synchronize the full states of the quadrotor with the time-varying states of the perching surface (Sec. \ref{subsec:Flexible Terminal Transformation}).
\subsection{Dynamic Model and Differential Flatness}
\label{subsec:dynamic model}
In this paper, we use the simplified dynamics proposed by \cite{mueller2015computationally} for a quadrotor, whose configuration is defined by its translation $\mathbf p = (p_x, p_y, p_z)^T \in \mathbb R^3$ and rotation $\mathbf R \in \mathrm{SO}(3)$. Translational motion depends on the gravitational acceleration $\bar g$ as well as the thrust $\tilde f$. Rotational motion takes the body rate $\bm \omega \in \mathbb R^3$ as input. The simplified model is written as 
\begin{equation}
  \begin{cases}
  \bm \tau = \tilde f \mathbf R \mathbf e_3 / m, \\
    \ddot{\mathbf p} = \bm \tau - \bar{g} \mathbf e_3, \\
    \dot{\mathbf R} = \mathbf R \hat{\bm \omega},
  \end{cases}
\end{equation}
where $\bm \tau$ denotes the mass-normalized net thrust, $\mathbf e_i$ is the $i$-th column of $\mathbf I_3$, such as $\mathbf e_3  = [0,0,1]^T$, and $\hat \cdot$ is the skew-symmetric matrix form of the vector cross product.

Exploiting the differential flatness property of the quadrotor, the state and input variables of quadrotors can be parameterized by finite derivatives of flat outputs \cite{mueller2015computationally}. The flat output of quadrotors is 
\begin{equation}
    \mathbf z = (p_x, p_y, p_z, \psi)^T \in \mathbb{R}^3 \times \operatorname{SO}(2),
\end{equation}
where $\psi \in SO(2)$ is the Euler-yaw angle.
We further define the flat outputs and their derivatives $\mathbf z^{[s-1]} \in \mathbb{R}^{ms}$ as
\begin{equation}
    \mathbf z^{[s-1]} \coloneqq (\mathbf z^T, \dot{\mathbf z}^T,...,{\mathbf z^{(s-1)}}^T)^T.
\end{equation}
This makes it possible for us to optimize a trajectory $\mathbf z(t):[0, T] \mapsto \mathbb{R}^m$ in the low-dimension flat-output space.

\subsection{Target Prediction Model}
\label{subsec:Estimation model}
Given measurements on the target position $(\varrho_x , \varrho_y , \varrho_z)$, we aim to estimate the full state of the target vehicle. 
Estimating high-order states will amplify the observation error from the position measurement noise.
Therefore, we adopt the constant turn rate and velocity (CTRV) model \cite{huang2022survey} for state estimation, and the states of the target vehicle are written as
\begin{equation}
    \boldsymbol{\chi}(t)=\left(\varrho_x , \varrho_y , \varrho_z, \theta , v_h , v_v , \omega_{\varrho}
    \right)^T,
\end{equation}
where $\theta$ is the heading angle of the vehicle, $\omega_{\varrho}$ is the angular velocity, $v_h$ is the horizontal velocity, and $v_v$ is the vertical velocity, which is usually small and used to estimate uphill and downhill motion.
Then we use EKF with this model for the state estimation.

To combine environmental information for target prediction, given the estimated current states, we use a kinodynamic motion primitive method considering collision. Given a prediction duration, by expanding motion primitives using the constant turn rate and acceleration (CTRA) model \cite{huang2022survey} under different control inputs with a time step $\Delta t$, we choose the collision-free one with the least control effort. Expanding until the prediction duration is reached, we obtain the predicted trajectory $\bm \varrho(t) \in \mathbb{R}^3$ with yaw angle $\theta (t)\in \operatorname{SO}(2)$ and a series of corresponding discrete target positions denoted as
\begin{equation}
\label{target predict path}
    \Psi = \left\{ \bm \varrho_k = \bm \varrho(k\Delta t) \mid k =\left[ 0, 1, \dots, M\right]   \right\}.
\end{equation}

\subsection{Geometric Model}
\label{subsec:Geometric model}
\begin{figure}[ht]
	\begin{center}
		\includegraphics[width=1.0\columnwidth]{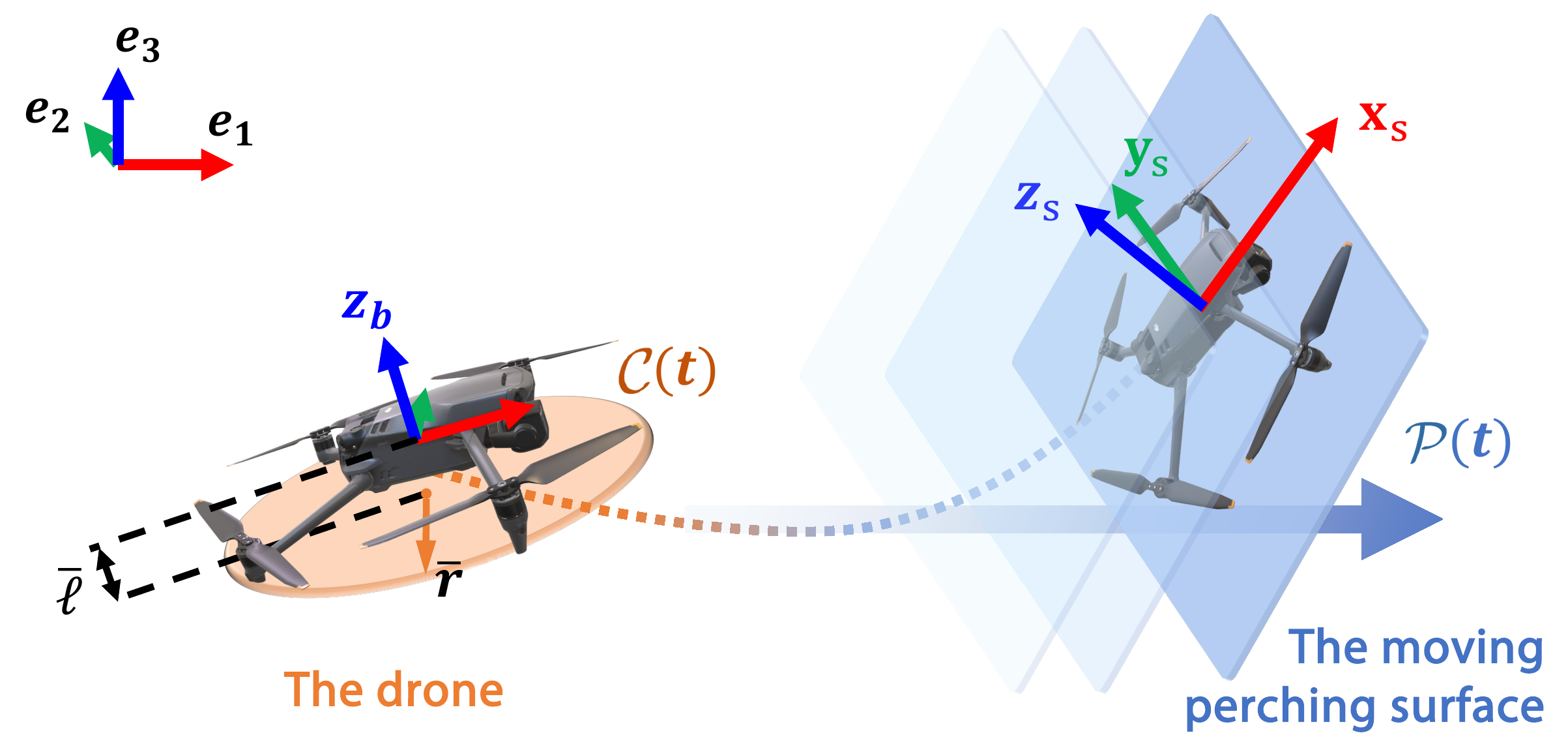}
	\end{center}
	\caption{
        \label{fig:model}
        Illustration of the geometry model. The quadrotor is modeled as a disc $\mathcal C$, and the perching surface is modeled as a half-space $\mathcal P$.
	}
\vspace{-0.5cm}
\end{figure}

To avoid collision with the perching surface, we model the underside of a symmetric quadrotor as a disc, shown in Fig.~\ref{fig:model}, denoted by
\begin{equation}
  \label{eq:C(t)}
  \mathcal C = \left\{\mathbf x = \mathbf R \mathbf B \bar r \mathbf u + \mathbf o ~\Big|~ \norm{\mathbf u} \leq 1, \mathbf u \in \mathbb R_2, \mathbf x \in \mathbb R_3 \right\},
\end{equation}
where $\mathbf B = \left(\mathbf e_1, \mathbf e_2\right) \in \mathbb R_{3\times2}$ and $\bar r$ denotes the radius of the disc. We use $\bar \ell$ to denote the thickness of the centroid of the quadrotor to the bottom. Thus, the center of the disc $\mathbf o = \mathbf p - \bar \ell \mathbf z_b$, where $\mathbf z_b$ is the normal vector corresponding to the z-axis of the body frame.

We assume that the estimated position and yaw angle of the moving platform is denoted as $\bm \varrho$ and $\theta$, and the normal vectors corresponding to the perching surface frame are denoted by $\mathbf x_s,\mathbf y_s$, and $\mathbf z_s$. Therefore, the feasible half-space divided by the perching surface is written as
\begin{equation}\\
  \label{eq:P(t)}
  \mathcal P = \left\{ \mathbf h^T \mathbf x \leq b ~\Big|~ \mathbf x \in \mathbb R_3 \right\},
\end{equation}
where $\mathbf h = - \mathbf z_s$, $ b = \mathbf h^T\bm \varrho$, and $\mathbf{z}_s=\mathbf{R}_z\left(\theta \right)\mathbf{\bar z}_s$, where $\mathbf{R}_z(\theta) \in \mathrm{SO}(3)$ represents the basic rotation by angle $\theta$ around ego z-axis, and $\mathbf{\bar z}_s$ is the current normal vector obtained by detection. The other normal vectors can be similarly calculated.

\section{Visibility-aware Tracking Planning}
\label{sec:Visibility-aware Tracking Metrics}
    As analyzed in Sec.~\ref{sec:intro}, planning with visibility consideration is essential for conducting long-term stable target tracking.
    To this end, we define several metrics to comprehensively model visibility in this section.
    Observing typical tracking failures, we can summarize that they are mainly due to observation distance (OD), observation angle (OA), and occlusion effect (OE).
    In this section, we explicitly model the above OD, OA, and OE metrics, design their penalty functions, and derive corresponding gradients for trajectory optimization. 
    In what follows, the drone's position and yaw angle are denoted as $\mathbf{p} \in \mathbb{R}^{3}$ and $\psi \in SO(2)$, with the target position $\bm \varrho \in \mathbb{R}^{3}$. 
 
\subsection{Observation Distance Restriction}
A target is expected to be observed in a proper distance range from a drone. 
Since target ground platforms usually have more aggressive horizontal motion, we separately constrain the horizontal ($\delta^h$) and vertical ($\delta^v$) components of the tracking distance with different margins: 
	\begin{equation}\label{equ:do}
		d_{\ell}^{*} \le \delta^{*} \le d_u^{*}, *=\{h,v\},
	\end{equation}
	where the $d_{\ell}$ and $d_u$ are the lower and upper bounds of the optimal distance of observation. 
    To make this metric analytically differentiable, we design penalty functions for the OD constraint:
    \begin{subequations}
    \label{equ:do_cost}
        \begin{align}
          \mathcal J_{OD^v} &= g(d_{\ell}^v - \delta^v) + g(\delta^v-d_u^v),\\
          \mathcal J_{OD^h} &= g\left(d_{\ell}^h - \delta^h\right) + \mathcal L_\mu \left(\delta^h - d_u^h\right),
        \end{align}
    \end{subequations}
    where $g(x)=\max\left( x,  0\right)^3$ and $\mathcal L_\mu(\cdot)$ is a $C^2$-smoothing linear penalty function denoted by
    \begin{align}
      \mathcal L_\mu(x) = 
      \begin{cases}
        0,                    & x \leq 0,       \\
        (\mu - x/2)(x/\mu)^3, & 0 < x \leq \mu, \\
        x - \mu/2,            & x > \mu.
      \end{cases}
    \end{align}
    When the horizontal distance surpasses the upper bound $\mu$, we apply a linearly increasing penalty function $L_\mu(\cdot)$.
    We choose a more severe penalty $g(\cdot)$ to prevent the drone from getting the target too close and possibly causing a collision.
    
	\subsection{Observation Angle Restriction}
	In order to keep the observation angle straight toward the target, the expected yaw angle $\psi_{e}$ is defined as
	\begin{equation}\label{equ:ao}
		\psi_{e}(\mathbf{p}, \bm \varrho) = {\rm atan2} \left(\mathbf e_2^T (\bm \varrho-\mathbf{p}), \mathbf e_1^T (\bm \varrho-\mathbf{p}) \right), 
	\end{equation}
	where $\mathbf e_i$ is the $i$-th column of $\mathbf I_3$.
	The cost of this term is written as
	\begin{equation}\label{equ:ao_cost}
        \mathcal J_{OA} = (\psi - \psi_{e})^2.
	\end{equation}
	Note that the drone could adjust both its position and yaw angle to reach $\psi_{e}$. The gradient of $\mathcal J_{OA}$ affects both $\mathbf {p}$ and $\psi$. Here we present the gradient w.r.t. $\mathbf {p}$ written as~$\footnote{Note that we adopt the denominator layout for gradients in this paper.}$
	\begin{equation}
		\frac{\partial{\mathcal J_{OA}}}{\partial{\mathbf{p}}} = \left[\frac{\partial{\mathcal J_{OA}}}{\partial{p_x}},\frac{\partial{\mathcal J_{OA}}}{\partial{p_y}},0  \right]^T,
	\end{equation}
	where
	\begin{equation}
		\frac{\partial{\mathcal J_{OA}}}{\partial{p_x}} = \frac{2\left(\psi - \psi_{e}\right)}{(\boldsymbol e_1^T\mathbf{(p-\bm \varrho)})^2 + (\boldsymbol e_2^T\mathbf{(p-\bm \varrho)})^2} \cdot \boldsymbol e_2^T\mathbf{(p-\bm \varrho)}.
	\end{equation}
	$\partial{\mathcal J_{OA}}/\partial{p_y}$ can be calculated similarly.

	\subsection{Elastic Occlusion Effect Avoidance}	
	\label{sec:OE}
    Complex environments increase the possibility of occlusion during tracking, easily causing target loss.
    Considering the conical FoV and arbitrarily shaped obstacles, we formulate the OE metric as follows.
	\begin{figure}[t]
		\centering
		\includegraphics[width=1\linewidth]{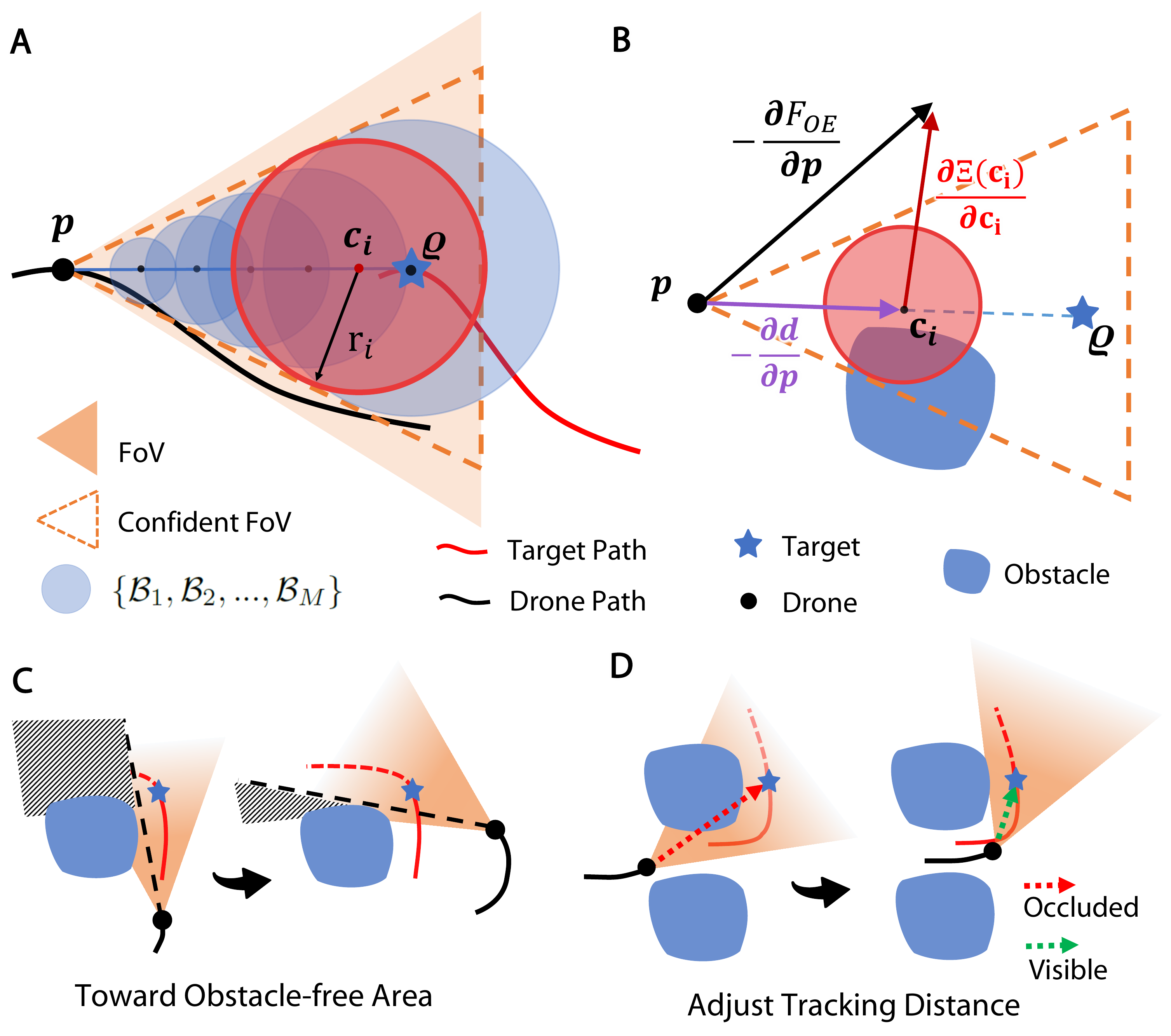}
        \vspace{-0.3cm}
		\caption{
			\textbf{A}. Illustration of the OE metric. The drone at position $\mathbf{p}$ observes the target $\bm \varrho$. A sequence of ball-shaped areas in blue are used to approximate the confident FoV expected to be obstacle-free. 
            \textbf{B}. An example of the two decomposition parts of the negative $F_{OE}$'s gradient, driving the drone to move towards obstacle-free areas and adjust the tracking distance elastically.
            \textbf{C}. The drone moves towards obstacle-free areas for less occlusion probability. \textbf{D}. The drone adjusts the tracking distance in obstacle-rich areas for less occlusion probability.
		}
		\label{fig:visi_metric}
		\vspace{-0.5cm}
	\end{figure}
	As shown in Fig.~\ref{fig:visi_metric}A, the \textit{confident FoV} embedded in the real FoV represents the region that is expected obstacle-free around the line of sight.
    To formulate analytically, we approximate the confident FoV with a sequence of ball-shaped areas $\{ \mathcal{B}_1,\mathcal{B}_2,...,\mathcal{B}_M \}$, where $M$ is the number of the areas. For each ball, its center $\mathbf c_i$ and radius $r_i$ are calculated by
	\begin{align}\label{eq:pi}
		\mathbf c_i &= \mathbf{p} + \lambda_i(\bm \varrho-\mathbf{p}), \\
		r_i &= \rho \cdot \lambda_i \cdot \norm{\mathbf{p}-\bm \varrho}, \label{eq:r_i}
	\end{align}
	where $\lambda_i = i/M \in [0, 1]$, and $\rho$ is a constant determined by the size of the confident FoV.
	Then we guarantee OE analytically by forcing constraint
	\begin{equation}
		\label{eq:oe}
		r_i < \Xi(\mathbf c_i),
	\end{equation}
    for each ball-shaped area,
	where $\Xi(\mathbf c_i): \mathbb R^3 \rightarrow \mathbb{R}$ is the distance between $\mathbf c_i$ and its closest obstacle, which is obtained from Euclidean Signed Distance Field (ESDF).
	Then, the OE cost is concisely written as
    \begin{subequations}
    \label{equ:eo_cost}
        \begin{align}
            \mathcal J_{OE} &= \sum_{i=1}^{M}g(F_{OE}(\mathbf{p})),\\
            F_{OE}(\mathbf{p}) &=  r_{i} - \Xi(\mathbf c_{i}).
        \end{align}
    \end{subequations}
	The gradient of $F_{OE}$ can be written as
	\begin{equation}
		\begin{aligned}
			\frac{\partial{F_{OE}}}{\partial{\mathbf{p}}} = & 
			\rho  \lambda_i \cdot \frac{\partial {d}}{ \partial \mathbf{p}} 
			-(1-\lambda_i) \frac{\partial{\Xi(\mathbf{c}_{i})}}{\partial{\mathbf{c}_{i}}},
		\end{aligned}
	\end{equation}
	where ${d}=\norm{\mathbf{p}-\bm \varrho}$, and $\partial\Xi(\mathbf c_{i})/\partial\mathbf{c}_{i}$ is obtained from ESDF. 
 
    To further analyze the metric, an example of the gradient of $F_{OE}$ is presented in Fig.~\ref{fig:visi_metric}B. 
    As the two directions of the decomposition parts of $-{\partial{F_{OE}}}/{\partial{\mathbf{p}}}$ shown, the OE metric can not only make the drone move towards obstacle-free areas (Fig.~\ref{fig:visi_metric}.C) but also adjust the tracking distance elastically according to the environment (Fig.~\ref{fig:visi_metric}.D). 
    Both of the two actions can prevent occlusion, especially when a target moves in obstacle-rich regions shown in Fig.~\ref{fig:visi_metric}.D, with a small positional adjustment margin, getting closer to the target is a reasonable way. 
    Such characteristic is further demonstrated in simulation experiments in Sec.~\ref{subsec:case study}.

\section{Aggressive and Flexible Perching Planning}
\label{sec:Robust and Dynamic Perching Metrics}
To smoothly attach to the perching surface, it is essential to get a full-state alignment at the contact moment, with safety, dynamic feasibility, and continuous target observation guaranteed. To this end, in this section, we detail the specific constraints involving the above factors for aggressive and flexible perching planning. 
In what follows, the position trajectory of the drone and the target are denoted as $\mathbf{p} \in \mathbb{R}^3$ and $\bm \varrho \in \mathbb{R}^3 $, respectively. The attitude quaternion of the quadrotor is denoted as  $\mathbf{q} \in \mathcal{S}^3$.

\subsection{SE(3) Collision Avoidance}
Considering the geometry of a quadrotor, adjusting its attitude is critical for collision avoidance with the perching surface, necessitating SE(3) trajectory planning. 
Given the description of $\mathcal C$ (Eq.~\ref{eq:C(t)}) and $\mathcal P$ (Eq.~\ref{eq:P(t)}), to avoid the quadrotor having intersection of the surface, the SE(3) collision avoidance constraint is written as
\begin{align}
  \mathcal C \subset \mathcal P,
\end{align}
which is equivalent to
\begin{subequations}
  \begin{align}
    \mathbf h^T \left(\mathbf R \mathbf B \bar r \mathbf u + \mathbf o  \right) - b \leq 0,                   \\
    \sup_{\norm{\mathbf u} \leq 1} \mathbf h^T \left(\mathbf R \mathbf B \bar r \mathbf u \right) + \mathbf h^T \mathbf o- b \leq 0, \\
    \label{eq:F(t)}
    \bar r \norm{\mathbf B^T \mathbf R^T \mathbf h} + \mathbf h^T \mathbf o - b \leq 0.
  \end{align}
\end{subequations}
Since we are utilizing the flat output $ \mathbf z $ as the state representation, we should first recover $\mathbf R$ from $ \mathbf z $ before computing the constraint Eq.~\ref{eq:F(t)}.
Firstly, the mass-normalized net thrust can be directly obtained from the trajectory: 
\begin{equation}
\label{eq:tau}
    \bm \tau = \textbf{p}^{(2)} + \bar{g} \mathbf e_3.
\end{equation}
Then, since the z-axis of the body frame $\textbf{z}_b$ is aligned with the direction of the mass-normalized net thrust according to the simplified dynamics \cite{mueller2015computationally}, we can obtain
\begin{equation}
\label{eq:zb}
    \textbf{z}_b =\bm \tau / \norm{\bm \tau}_2.
\end{equation}
To recover $\mathbf R$ from $\textbf{z}_b$, denoting that $\textbf{z}_b = [a_z, b_z, c_z]^T$, we utilize an efficient mapping according to  Hopf fibration \cite{watterson2020control}, which is written as
\begin{align}
  \textbf{q}_{a b c}=\frac{1}{\sqrt{2(c_z+1)}}\left(\begin{array}{c}c_z+1 \\-b_z \\a_z \\0\end{array}\right).
\end{align}
where the unit quaternion $\textbf{q}_{a b c}$ satisfies $\mathbf R\left(\textbf{q}_{a b c}\right) \mathbf e_{3} = \textbf{z}_b$.
Since this collision avoidance metric is independent of the yaw angle $\psi$, the rotation of the quadrotor can be obtained by $\mathbf R=\mathbf R\left(\textbf{q}_{a b c}\right)$.
Adopting this differential flatness with Hopf fibration can reduce computation and reduce the singularities to the best case possible.
Finally, we can easily calculate $\mathbf{B}^{T} \mathbf R^T$ in Eq.~\ref{eq:F(t)} through $\textbf{z}_b$:
\begin{equation}
\label{eq:BTRT}
  \mathbf{B}^{T} \mathbf R^T=
  \begin{pmatrix}
1-\frac{a_z^{2}}{c_z+1} & -\frac{a_z b_z}{c_z+1} & -a_z \\ -\frac{a_z b_z}{c_z+1} & 1-\frac{b_z^{2}}{c_z+1} & -b_z
  \end{pmatrix}.
\end{equation}
However, the constraint Eq.~\ref{eq:F(t)} should only be activated when $\norm{\mathbf p - \bm \varrho} \leq \bar d$, that is, the drone is close to the platform.
This constitutes a mixed-integer nonlinear programming problem.
Here we design a smoothed logistic function $\mathcal L_\epsilon(\cdot)$ to incorporate the integer variable into our NLP, which is denoted by
\begin{align}
\label{eq:L01}
\mathcal L_\epsilon(x) = 
\begin{cases}
0,                                     & x \leq -\epsilon,     \\
\frac{1}{2\epsilon^4}(x+\epsilon)^3(\epsilon-x), & -\epsilon < x \leq 0, \\
\frac{1}{2\epsilon^4}(x-\epsilon)^3(\epsilon+x) + 1, & 0 < x \leq \epsilon,  \\
1,                                     & x > \epsilon,
\end{cases}
\end{align}
where $\epsilon$ is a tunable positive parameter. 
Then the penalty function of collision avoidance can be written as
\begin{align}
\label{SE3 cost function}
	\mathcal J_c &= \mathcal L_\epsilon \left(\mathcal F_{\bar{d}} \right)
	\cdot \mathcal L_\mu \left(\mathcal F_c \right),
\end{align}
where $\mathcal F_{\bar{d}}$ and $\mathcal F_c$ are denoted by
\begin{subequations}
	\begin{align}
    	\mathcal F_{\bar{d}} &= {\bar d}{}^2 - \norm{\textbf{p} - \bm{\varrho}}^2,\\
		\mathcal F_c &= \bar r \norm{\mathbf B^T \mathbf R ^T \mathbf h} + \mathbf h^T \mathbf o- b.
	\end{align}
\end{subequations}

Additionally, to limit the drone within a safe relative height range and prevent the drone from hitting the ground, we define a relative height cost $\mathcal J_{\Delta z}$ written as
\begin{align}
\label{eq:relative z cost function}
  \mathcal J_{\Delta z} = 
  \mathcal L_\mu \left({z}_{min} - {\Delta z} \right) + \mathcal L_\mu \left({\Delta z} - {z}_{max}\right),
\end{align}
where relative height $\Delta z = \mathbf e_3^T (\textbf{p} - \bm{\varrho})$, ${z}_{min}$ and ${z}_{max}$ are the expected minimum and maximum relative heights.

\subsection{Image Space Pixel-level Perception Enhancement}
For accurate perching, high-quality observation is a fundamental factor.
With a visual tag that remarks the perching position, the quadrotor is expected to keep it central in the camera image. 
Denoting the tag position that coincided with the perching position in the world frame as ${}^w \bm{\varrho}$, we first transform it to the camera frame:
\begin{subequations}
  \begin{align}
    {}^b\bm{\varrho} &= \mathbf R({}^b_w \textbf{q} )({}^w \bm{\varrho}  - \textbf{p}),\\
    {}^c \bm{\varrho} &= \mathbf R({}^c_b \textbf{q})({}^b\bm{\varrho} - {}^c_b \textbf{t}),
  \end{align}
\end{subequations}
where we denote index $w$ as world frame, $c$ as camera frame and $b$ as body frame. ${}^c_b \textbf{q}$ and ${}^c_b \textbf{t}$ are the rotation quaternion and the translation between the body frame and the camera frame, respectively. ${}^b_w \textbf{q}$ is the rotation quaternion between the world frame and the body frame.
\begin{figure}[t]
    \begin{center}
        \includegraphics[width=0.93\columnwidth]{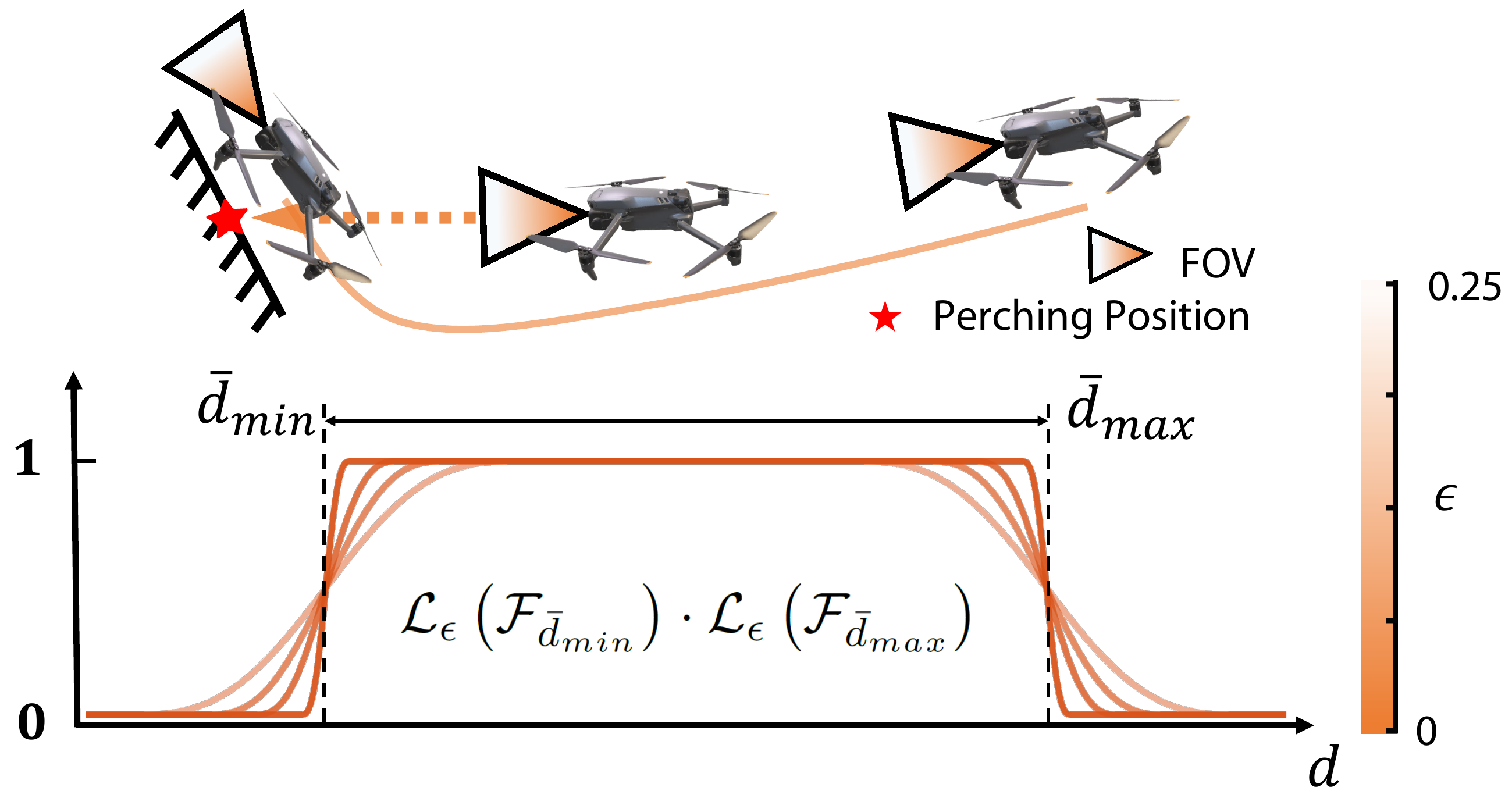}
    \end{center}
    \vspace{-0.5cm}
    \caption{Illustration of the perception distance range selection function with a front camera. The perception metric is only activated within the setting range $\left[\bar{d}_{min},\bar{d}_{max} \right]$. Overly constraining at a close relative distance can lead to violations of terminal attitude constraints.}
    \vspace{-0.5cm}
    \label{fig:percept_range}
\end{figure}
\begin{figure}[b]
    \begin{center}
        \includegraphics[width=1.0\columnwidth]{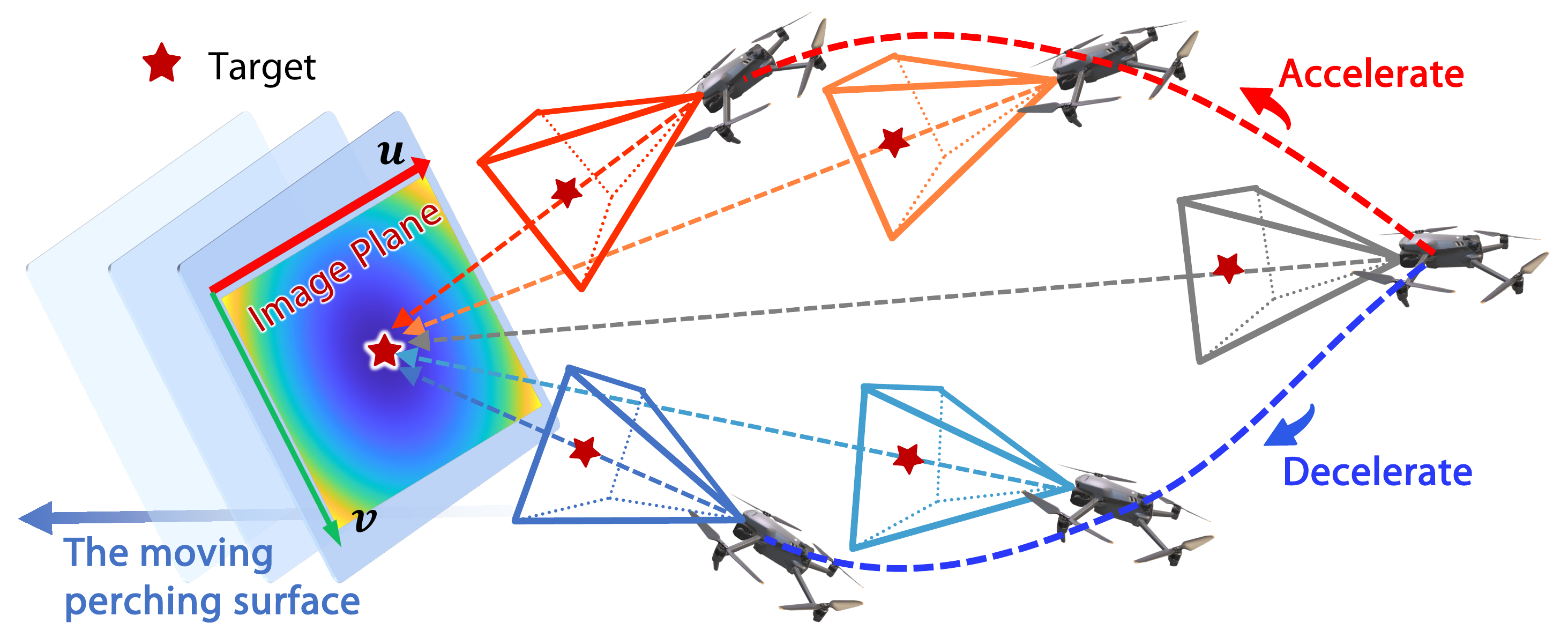}
    \end{center}
    \vspace{-0.2cm}
    \caption{For perception enhancement, the target is expected to be locked centrally in the image space. Since the motion and view attitude of a quadrotor are coupled, during acceleration or deceleration, the quadrotor needs to jointly adjust its attitude and position as shown to guarantee the perception quality.}
    \label{fig:project}
\end{figure}
Using the pinhole camera model, the tag position in the camera frame, with the center of the image as the origin, can be written as
\begin{equation}
    \left[
        \begin{array}{l}
            u_c \\ 
            v_c
        \end{array}
    \right]=
    \left[
        \begin{array}{l}
            f_x {}^{c} \bm{\varrho}_x / {}^{c} \bm{\varrho}_z \\
            f_y {}^{c} \bm{\varrho}_y / {}^{c} \bm{\varrho}_z
        \end{array}
    \right].
\end{equation}
where $f_x, f_y$ are the known camera intrinsics. 
Excessive constraints at a far distance limit the spatial freedom of trajectories, while overly constraining at a close distance might lead to violations of terminal attitude constraints, as shown in Fig.~\ref{fig:percept_range}.
Therefore, perception constraint should be activated only when the drone is within $\left[\bar{d}_{min},\bar{d}_{max} \right]$ distance range from the perching surface, introducing mix-integer programming problem.
We still resort to $\mathcal L_\epsilon(\cdot)$ defined in Eq.~\ref{eq:L01} to incorporate the integer variable.
Then the penalty function of tag perception is written as
\begin{align}
\label{perception cost function}
	\mathcal J_p &= \mathcal L_\epsilon \left(\mathcal{F}_{\bar{d}_{min}} \right) \cdot \mathcal L_\epsilon \left(\mathcal{F}_{\bar{d}_{max}}\right) \cdot \mathcal F_p,
\end{align}
where $\mathcal F_p$, $\mathcal{F}_{\bar{d}_{min}}$, and $\mathcal{F}_{\bar{d}_{max}}$ are denoted by
\begin{subequations}
\begin{align}
	\mathcal F_p &= u_c^2 + v_c^2,\\
    \mathcal{F}_{\bar{d}_{min}}&=\|\mathbf p-{}^w \bm{\varrho}\|^2-\vec{d}_{min}^2,\\
    \mathcal{F}_{\bar{d}_{max}}&=\vec{d}_{max}^2 - \|\mathbf p-{}^w \bm{\varrho}\|^2.
\end{align}
\label{eq:F_perception}
\end{subequations}
As $\mathcal F_p$ implies, the drone position $\textbf{p}$ and attitude $\textbf{q}$ jointly affect the perception quality. But the coupled attitude and motion of the quadrotor further hinder perception enhancement, as shown in Fig.~\ref{fig:project}. 
To address the problem, we provide the gradients of $\mathcal F_p$ w.r.t. $\textbf{p}$ and $\textbf{q}$ here, and optimize the trajectory in quadrotor flat-output space in latter Sec.~\ref{subsec:traj opt}.
The gradient w.r.t. $\textbf{p}$ is written as
\begin{align}
	\frac{\partial \mathcal F_p}{\partial \textbf{p}}  =
	\frac{\partial {}^c \bm{\varrho}}{\partial \textbf{p}}
	\frac{\partial \mathcal F_p}{\partial {}^c \bm{\varrho}},
\end{align}
where $\frac{\partial {}^c \bm{\varrho}}{\partial \textbf{p}} = -({}^c_b \mathbf R \cdot {}^b_w \mathbf R)^T$.
Moreover, the gradient w.r.t. the drone's attitude quaternion can be written as
\begin{align}
	\frac{\partial \mathcal F_p}{\partial {}^w_b \textbf{q}}=
	(
	\frac{\partial {}^b \bm{\varrho}}{\partial {}^b_w \textbf{q}}
	\frac{\partial {}^c \bm{\varrho}}{\partial {}^b \bm{\varrho}}
	\frac{\partial \mathcal F_p}{\partial {}^c \bm{\varrho}})^{-1},
\end{align}
where $\frac{\partial {}^b \bm{\varrho}}{\partial {}^b_w \textbf{q}}$ is the jacobian matrix of a quaternion rotation w.r.t the quaternion, $\frac{\partial {}^c \bm{\varrho}}{\partial {}^b \bm{\varrho}} = {}^b_c \mathbf R$, and $(\cdot)^{-1}$ is the inversion of quaternion. 
The gradient w.r.t. attitude quaternion can be further transformed to gradients w.r.t. the flat-output states of the drone according to \cite{watterson2020control}.


\subsection{Actuator Constraints}
To ensure dynamic feasibility during agile flight, we constrain the quadrotor's thrust and angular velocity within a reasonable range.
Firstly, the mass-normalized net thrust calculated by Eq.~\ref{eq:tau} is bounded as 
\begin{align}
    \tau_{min} \leq \norm{\bm \tau} \leq \tau_{max},
\end{align}
which can be constrained by constructing such a penalty function
\begin{align}
\label{tau cost function}
  \mathcal J_{\tau}  = \mathcal L_\mu (\norm{\bm \tau}^2 -  \tau_{max}^2 ) + \mathcal L_\mu ( \tau_{min}^2 - \norm{\bm \tau}^2 ).
\end{align}
Secondly, the limitation of body rate $\bm \omega= [\omega_x,\omega_y, \omega_z]^T = \mathbf R^T \dot{\mathbf R}$ can also be constrained by penalty function
\begin{align}
\label{omega cost function}
  \mathcal J_{\omega}  & = \mathcal L_\mu \left( \norm{\bm \omega_{xy}}^2 - \omega_{xy,max}^2 \right)\\
  &+ \mathcal L_\mu \left( \norm{\omega_z}^2 - \omega_{z,max}^2 \right),
\end{align}
where $\bm \omega_{xy} = [\omega_x,\omega_y]^T$, $\omega_{xy,max}$ is the x-y body axis maximum angular velocity and $\omega_{z,max}$ is the z-axis part.
To reduce singularities and simplify calculation, we enforce the constraint with Hopf fibration angle velocity decompose~\cite{watterson2020control}:
\begin{align}
    &\norm{\bm \omega_{xy}}^2 = \omega_x^2 + \omega_y^2 = \norm{\dot{\mathbf z}_b},
\end{align}
where $\dot{\mathbf z}_b$ is calculated by
\begin{align}
\label{eq: zb_dot}
    \dot{\mathbf z}_b = f_{\cal DN}\left(\bm \tau \right) \mathbf{p}^{(3)},
\end{align} 
where $\mathbf{p}^{(3)}$ represents the jerk of the drone on the trajectory, and $f_{\cal DN}(\cdot)$ is given by
\begin{align}
\label{eq:fdn}
  f_{\cal DN}(x) = \left(\mathbf{I}_3 - \frac{xx^T}{x^Tx}\right) / \norm{x}_2.
\end{align}
Since the yaw angle $\psi$ changes a little while perching, we approximately adopt $\omega_z=\dot{\psi}$.

 

\subsection{Flexible Terminal Adjustment}
\label{subsec:Flexible Terminal Transformation}

\begin{figure}[ht]
    \begin{center}
         \includegraphics[width=1.0\columnwidth]{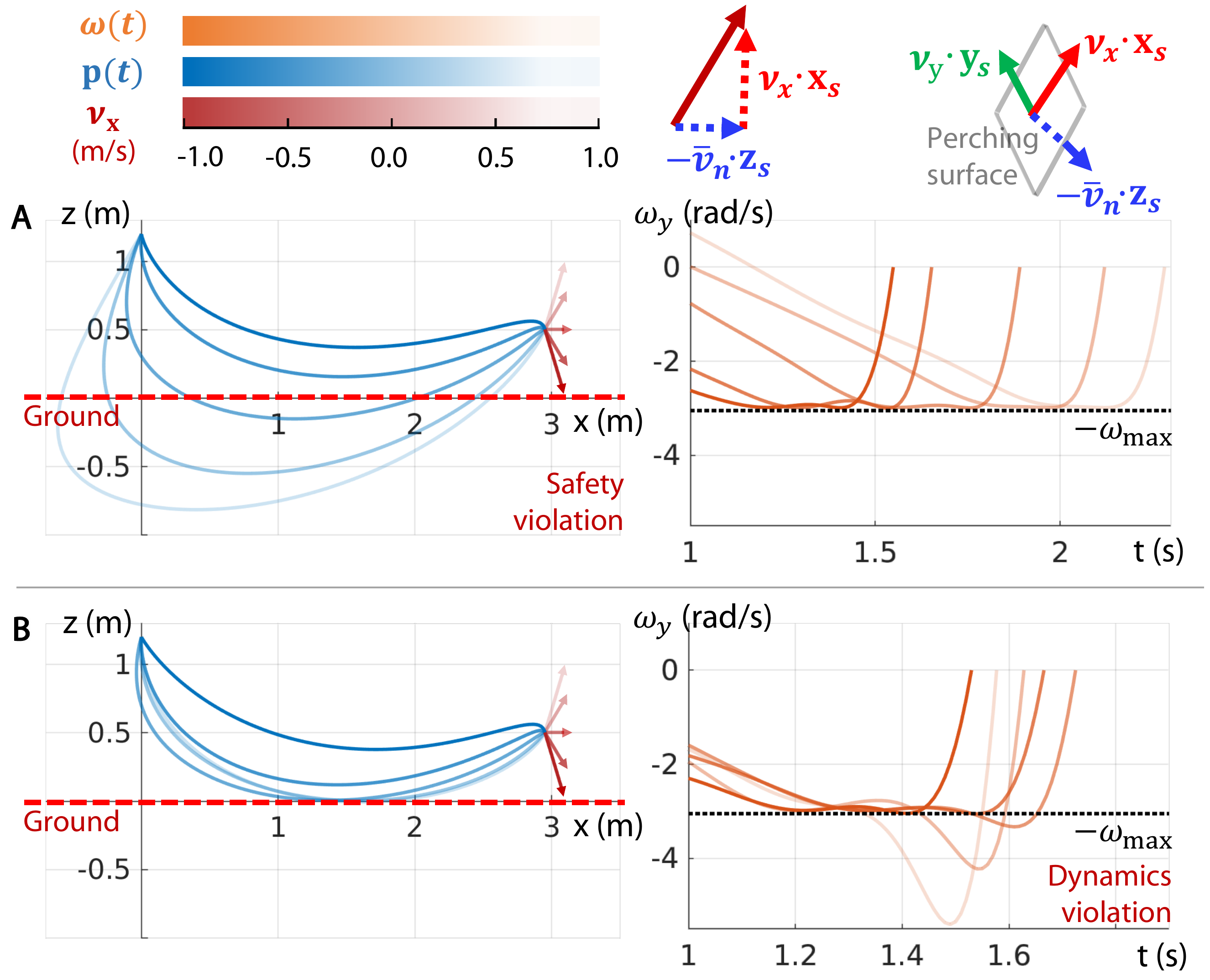}
    \end{center}
    \caption{The optimized perching trajectories for $1.5\, \text{rad}$ terminal attitude with different tangential relative speeds along $\mathbf{x}_s$: $\nu_{x}=\{-1.0,-0.5,0.0,0.5,1.0\} \,(\text{m/s})$. This instance shows the relative end speed constraint can conflict with safety and actuator constraints. \textbf{A}. With a hard actuator constraint, trajectories with $\nu_{x}=\{0.0,0.5,1.0\} \,(\text{m/s})$ violate the safety constraint and hit the ground ($z=0$). \textbf{B}. With a hard safety constraint, trajectories with  $\nu_{x}=\{0.0,0.5,1.0\} \,(\text{m/s})$ violate the actuator constraint $\omega_{max}=3\, \text{rad/s}$. Hence, it is necessary to optimize the tangential relative velocity $\boldsymbol \nu$ to simultaneously satisfy safety and actuator constraints.}\label{fig:vt}
\end{figure}

To attach to the moving perching platform quickly and smoothly, the quadrotor should align its terminal states with the perching surface at a proper moment.
During trajectory optimization, the temporal profile is affected by the above dynamics, safety constraints, and the time regularization introduced later in Sec.~\ref{subsec:traj opt}, determining the contact moment.
As the flight duration changes, the desired terminal states of the quadrotor vary due to the motion of the perching surface.
To flexibly synchronize the quadrotor's position, velocity, and attitude with the time-varying states of the perching surface, we formulate the terminal constraints of the drone trajectory as a function of the predicted target trajectory as follows.

At the contact moment, which corresponds to the end of the perching trajectory, the quadrotor should exactly coincide with the perching pose:
\begin{align}
    \textbf{p}(T)=\bm{\varrho}(T) + \bar l \mathbf z_s(T),
\end{align}
where $T$ is the trajectory duration.
Additionally, we expect the heading of the quadrotor to coincide with the estimated target's heading:
\begin{align}
    \psi(T)=\theta(T).
\end{align}
Ideally, the drone needs to stay relatively still from the target:
\begin{align}
    \label{eq:landing relative vel}
    \textbf{p}^{(1)}(T) = \bm \varrho^{(1)}(T) - \bar{v}_n\mathbf z_s(T),
\end{align}
where $\mathbf{p}^{(1)}$ and $\bm{\varrho}^{(1)}$ are the velocity trajectory of the drone and the target, $\bar{v}_n$ is a preset small normal relative speed that helps the drone to stick to the perching surface at the end.
However, the hard constraints of the relative end state may conflict with either the safety or the actuator constraints, as shown in Fig.~\ref{fig:vt}. The conflicting constraints will result in no feasible trajectories. Moreover, considering the observation error, the strict terminal velocity constraint may lead to attitude shaking.
Therefore, we relax the end velocity constraint by adding the tangential relative speed $\boldsymbol \nu=[ \nu_{x},\nu_{y}]^T \in \mathbb R^2$ as a new variable to be optimized into Eq.~\ref{eq:landing relative vel}:
\begin{subequations}
  \begin{align}
    \textbf{p}^{(1)}(T) &= \bm{\varrho}^{(1)}(T) + \boldsymbol{\delta}_v(T),\\
    \boldsymbol{\delta}_v(T) &=  \left[ \nu_{x}  \mathbf{x}_s(T) , \nu_{y}  \mathbf{y}_s(T), -\bar{v}_n\mathbf z_s(T) \right]^T,
  \end{align}
\end{subequations}
where $\nu_{x}$ and $\nu_{y}$ are the relative speed along the direction of $\mathbf{x}_s$ and $ \mathbf{y}_s$, respectively.
We minimize the tangential relative speed by introducing a regulation term
\begin{align}
\label{eq:relative vt}
	\mathcal J_{\nu} = \norm{\boldsymbol \nu}^2.
\end{align}
For attitude alignment, the z-axis of the drone body frame should coincide with the normal vector of the perching surface:
\begin{align}
\label{eq:perching zb}
    \mathbf z_b(T) = \mathbf z_s(T),
\end{align}
which is equivalent to
\begin{align}
\bm \tau(T) / \norm{\bm \tau(T)}_2 = \mathbf z_s(T).
\end{align}
To deal with the coupled motion and attitude of the quadrotor, we transform the attitude constraint to the final acceleration $\textbf{p}^{(2)}(T)$ in flat-out space.
Considering this constraint and the thrust limit for terminal states, we design a transformation:
\begin{subequations}
\begin{align}
	\textbf{p}^{(2)}(T) = \tau_{e} \cdot \mathbf z_s(T) - \bar g \mathbf e_3,\\
        \tau_{e} = \tau_{m} + \tau_{r} \cdot \sin(\tau_{f}).
\end{align}
\end{subequations}
where $\tau_{m} = (\tau_{max} + \tau_{min})/2$ and $\tau_{r} = (\tau_{max} - \tau_{min})/2$. 
By introducing a new variable $\tau_f \in \mathbb R$, the terminal thrust $\tau_{e}$ is limited within $[\tau_{min}, \tau_{max}]$ implicitly. 
The terminal attitude constraint is eliminated by this transformation.

As for terminal jerk, since it is related to the terminal angular velocity according to Eq.~\ref{eq: zb_dot}, we should set the final jerk $\mathbf p^{(3)}(T) = \mathbf 0$ to make the relative angular velocity of the robot small when it touches the surface.
Finally, the above terminal state constraints are collectively written as
\begin{align}
\label{eq:ternimal constraint}
    \mathbf z^{[s-1]}(T)=F(\bm \varrho(T)).
\end{align}

\section{Spatial-temporal Trajectory Optimization}
\label{sec:trajectory optimization}

\subsection{Occlusion-aware Path Finding}
To provide a reasonable initial path for tracking trajectory optimization, we design an occlusion-aware path-finding method that considers both tracking distance and occlusion.
Given the target predicted trajectory $\bm \varrho(t)$ and corresponding target series $\Psi = \left[ \bm \varrho_1,..., \bm \varrho_K \right]$ by Eq.~\ref{target predict path}, the path-finding method is aimed to offer a viewpoint series $\mathbf{V} = \left[ \mathbf{v}_1,..., \mathbf{v}_K \right]$ corresponding to $\Psi$ and a path connecting all viewpoints $\mathbf{V}$ for subsequent trajectory optimization. The whole tracking trajectory generation algorithm is listed as Alg.~\ref{alg1}.

\begin{algorithm}[t]
\caption{Visibility-aware Tracking Trajectory Generation}
\label{alg1}
\begin{algorithmic}[1]
	\Require target trajectory $\bm \varrho(t)$ and corresponding series $\Psi = \left[ \bm \varrho_1,..., \bm \varrho_K \right]$, current state\ $\mathbf{z}$, desired tracking distance $\bar{d}$, viewpoint series $\mathbf{V}$, tracking trajectory $\mathfrak{T}$;
    \State $\mathbf{V}$.clear; $Path$.clear;
    \State $\mathbf{v}_1 = \mathbf{z}$;
	\For{$i = 2$ to $K$}
    \State $Ray_{i-1} = \bm \varrho_{i-1} - \mathbf{v}_{i-1}$;
    \State $Traj_{i-1} \leftarrow \textbf{TrajSegment}(\bm \varrho(t),\bm \varrho_{i-1},\bm \varrho_{i})$;
    \For{$\mathbf{s}_j$ on $Ray_{i-1}$ and $Traj_{i-1}$}
    \If{\textbf{CheckRay}($\mathbf{s}_j$, $\bm \varrho_{i}$)}
    \State $\mathbf{s}_{res} = \mathbf{s}_j$;
    \State \textbf{break};
    \EndIf
    \EndFor
    \State $Dir_{i}= \mathbf{s}_{res}-\bm \varrho_{i}$;
	\State $\mathbf{v}_{i} \leftarrow$ \textbf{Extend(}$\mathbf{s}_{res},Dir_{i},\bar{d}$\textbf{)};
    \State $\mathbf{V}$.push\_back$(\mathbf{v}_{i}) $;
    \State $Path$.push\_back(\textbf{SearchPath}($\mathbf{v}_{i-1}, \mathbf{v}_{i}$));
	\EndFor
	\\$\mathfrak{T} \leftarrow $  \textbf{TrackingTrajOpt(}$Path$, $\Psi,\,\mathbf{V}$\textbf{)};
	\\\textbf{Return} $\mathfrak{T}$;
	\end{algorithmic}
\end{algorithm}

\begin{figure}[t]
	\begin{center}
		\includegraphics[width=0.93\columnwidth]{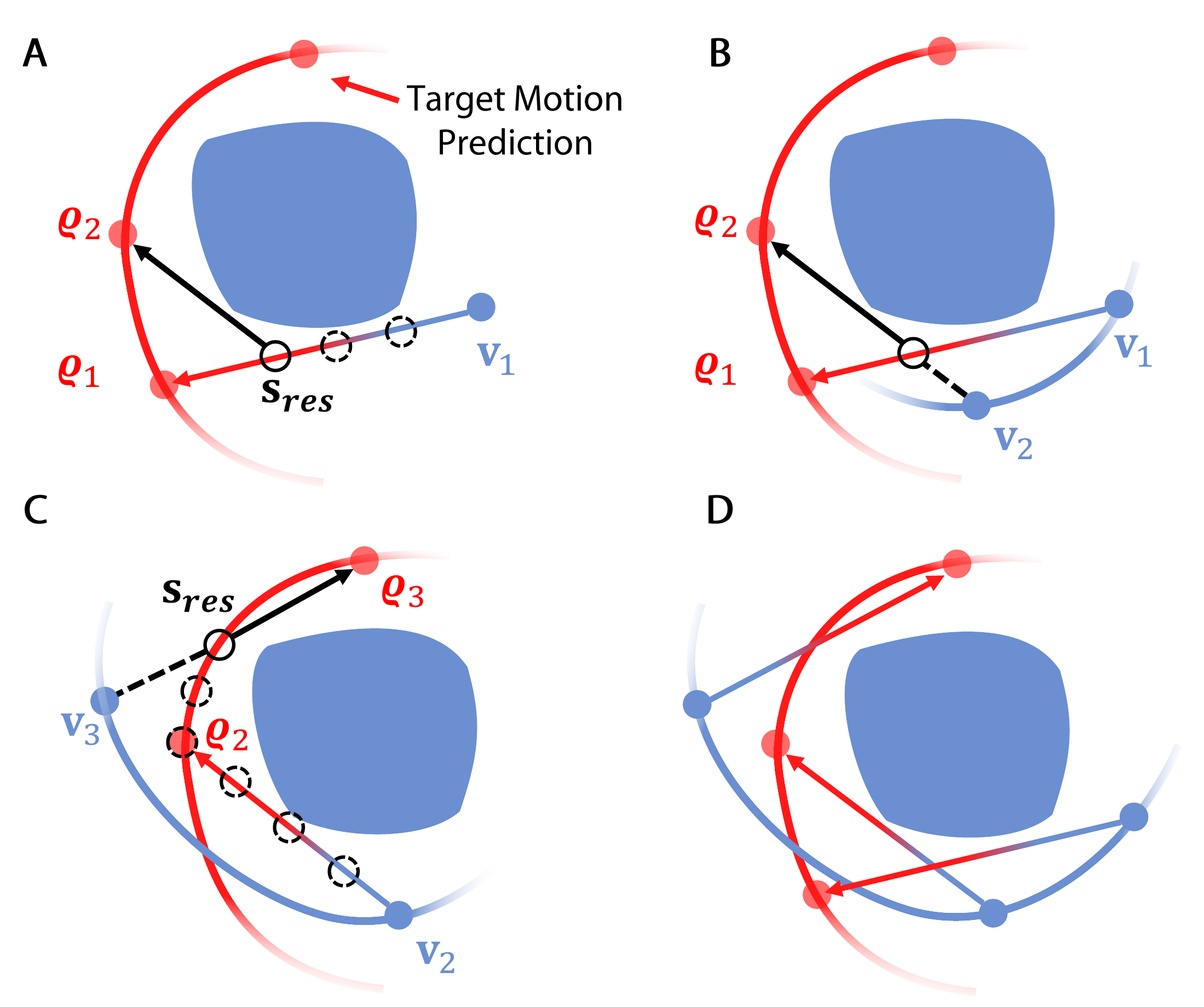}
	\end{center}
    \vspace{-0.5cm}
	\caption{
		\label{fig:front_end}
		Illustration of the occlusion-aware multi-goal path-finding method. 
	}
\end{figure}

Considering efficiency, we use the greedy method, decoupling the whole path-finding problem into smaller multi-goal path-searching problems. 
As an example shown in Fig.~\ref{fig:front_end}A ($i=2$ in this example), given the last viewpoint $\mathbf{v}_{i-1}$ and target point $\bm \varrho_{i-1}$, we first traverse the ray from $\mathbf{v}_{i-1}$ to $\bm \varrho_{i-1}$ to get sample points.
Then we check the line-of-sight from the sample point $\mathbf{s_j}$ to the current target point $\bm \varrho_{i}$ in function $\textbf{CheckRay()}$, and select the first no-occlusion one as $\mathbf{s}_{res}$ (Line 4-11). 
To keep the expected tracking distance, as shown in Fig.~\ref{fig:front_end}B, we then extend $\mathbf{s}_{res}$ along $\overline{\bm \varrho_i \mathbf{s}_{res}}$ in function $\textbf{Extend()}$, until the length of $\overline{\mathbf v_i \bm \varrho_i}$ equals to the expected tracking distance $\bar{d}$ or reach an obstacle (Line 12-13). 
If no valid sample point on the ray, we continue sample points on the target trajectory segment between $\bm \varrho_{i-1}$ to $\bm \varrho_{i}$, which is obtained by the function $\textbf{TrajSegment()}$.
An example is shown in Fig.~\ref{fig:front_end}C ($i=3$ in this example).
Thus, we obtain an occlusion-free viewpoint and simply use $\mathbf A^{\star}$ algorithm on the grid map to obtain the path from $\mathbf v_{i-1}$ to $\mathbf v_i$ in function $\textbf{SearchPath()}$ (Line 15). Subsequent viewpoints can be iteratively found. The found path and viewpoints set are used for later tracking trajectory optimization (Line 17).

\subsection{Trajectory Optimization Problem Formulation}
\label{subsec:traj opt}
Here we present the general trajectory optimization formulation, and the detailed design for specific stages of tracking and perching is described in Sec. \ref{subsec: tracking constraints} and Sec. \ref{subsec: perching constraints}, respectively.
We use $M$-piece polynomial to represent the piecewise flat-output trajectory $\mathbf z(t)$. $\textbf{T}=(T_1,\cdots,T_M)^T\in \mathbb{R}^M_{>0}$ are the durations of each piece, and $\textbf{c}=(\mathbf c_1^T,\cdots,\mathbf c_M^T)^T\in \mathbb{R}^{(N+1)\times 4M}$ are the coefficient matrices of each piece. Then, the $i^{th}$ piece is a $N$-degree polynomial
\begin{equation}
    \label{equ:i-th piece}
    \mathbf z_i(t)=\mathbf c_i^T \bm \beta(t),~~\forall{t}\in[0,T_i],
\end{equation}
where $\bm \beta(t)=[t^0,t^1,\cdots,t^N]^T$ is the natural basis.
The trajectory optimization problem can be formulated as follows:
\begin{subequations}
\label{eq:traj opt problem}
  \begin{align}
    \min_{\textbf{c}, \textbf{T}} & \label{eq:cost function}~\mathcal J_E = \int_{0}^{T} {\norm{\mathbf z^{(s)}(t)}^2} \df{t} + \rho T,                                                      \\
    s.t.~          & \label{eq:T>Tl} ~T \geq T_{\ell},\\
                   & \label{eq:init pos}~\mathbf z^{[s-1]}(0)=\bar{\mathbf{z}}_{0},                                                        \\
                   & \label{eq:final pos}~\mathbf z^{[s-1]}(T)=F(\bm \varrho(T)),                                                        \\
                   & \label{eq:user define G relative}~\mathcal G(\mathbf z^{[s-1]}(t),\bm \varrho(t)) \preceq \mathbf{0}, \forall t \in [0,T],\\
                   & \label{eq:user define G absolute}~\mathcal H(\mathbf z^{[s-1]}(t), \bm \varrho(t)) \preceq \mathbf{0}, \forall t \in \mathcal T,
  \end{align}
\end{subequations}
where $T = \sum_{i=1}^M T_i$, cost function Eq.~\ref{eq:cost function} trades off the smoothness and aggressiveness, and $\rho$ is the time regularization parameter. Eq.~\ref{eq:T>Tl} guarantees the entire duration greater than the lower bound $T_{\ell}$. Eq.~\ref{eq:init pos} and Eq.~\ref{eq:final pos} are the boundary conditions. Eq.~\ref{eq:init pos} guarantees the trajectory starting from initial state $\bar{\mathbf{z}}_{0} = \{\mathbf{z}_0,...,\mathbf{z}_0^{(s-1)} \}$. Eq.~\ref{eq:final pos}, which is explained in Sec.~\ref{subsec:Flexible Terminal Transformation}, constrains the final state determined by the predicted target trajectory. For different trajectory requirements, we design different additional constraints: Eq.~\ref{eq:user define G relative} are the inequality constraints that are continuously forced on the entire time, and Eq.~\ref{eq:user define G absolute} are the inequality constraints that are discretely forced on the specific time series $\mathcal T$.
The discrete-time constraints are designed to ensure consistency with the discrete results of the path-finding method.

To deform the spatial and temporal profiles of the trajectory during optimization efficiently, we adopt $\mathfrak{T}_{\mathrm{MINCO}}$ \cite{wang2022geometrically}, a state-of-the-art minimum control effort polynomial trajectory class. MINCO efficiently conducts spatiotemporal deformation of the M-piece flat-output trajectory $\mathbf z(t)$ by decoupling the space and time parameters with a linear-complexity mapping
\begin{equation}
    \label{equ:mapping}
    \mathbf z(t)=\mathcal{M}_{\textbf{m},\textbf{T}}(t),
\end{equation}
where $\textbf{T}\in \mathbb{R}^M_{>0}$ are the durations defined as aforemention and $\textbf{m}=(\textbf{m}_1,\cdots,\textbf{m}_{M-1})^T\in \mathbb{R}^{4\times(M-1)}$ are the adjacent intermediate points between connected pieces. 
The $s$-order $\mathfrak{T}^s_{\mathrm{MINCO}}$ consisting of $(2s-1)$-degree polynomials can represent an s-integrator chain dynamics system.
Furthermore, MINCO is advanced in converting $\{\textbf{m}, \textbf{T}\}$ to $\{\textbf{c}, \textbf{T}\}$ by the parameter mapping $\textbf{c} = C(\textbf{m}, \textbf{T})$ with linear
time and space complexity via \textit{Banded PLU Factorization}. Meanwhile, the gradients for $\{\textbf{c}, \textbf{T}\}$ are also propagated to MINCO parameters $\{\textbf{m}, \textbf{T}\}$ in linear time. We refer readers to \cite{wang2022geometrically} for more details.

To solve the continuous constrained optimization problem conveniently, we convert it to an unconstrained optimization problem. 
All kinds of equality constraints are implicitly satisfied by using the variable of MINCO for optimization.
To eliminate the inequality constraints Eq.~\ref{eq:user define G relative} - Eq.~\ref{eq:user define G absolute}, we use penalty function method introduced in Sec. \ref{Inequality Constraints Elimination}.
We adopt appropriate penalty weights and a fast post-optimization check for safety and dynamic feasibility to  prevent unreasonable inequality constraint violations.
The temporal constraint Eq.~\ref{eq:T>Tl} is eliminated by diffeomorphic variable substitution, which is introduced in Sec. \ref{Temporal Constraints Elimination}.  Finally, Eq.~\ref{eq:traj opt problem} is transfomed to an unconstrained optimization problem written as
\begin{equation}
\label{equ:unconstainted problem}
    \begin{aligned}
        \min_{
              \begin{scriptsize}
                \textbf{m},\textbf{T}
              \end{scriptsize}
             }  
          ~\mathcal J_E + \int_{0}^{T}  \mathcal{J}_{\mathcal G} dt + \sum_{t \in \mathcal T} \mathcal{J}_{\mathcal H} ,
    \end{aligned}
\end{equation}
where $\mathcal{J}_{\mathcal G}$ is the continuous-time penalty function corresponding to Eq.~\ref{eq:user define G relative}, and $\mathcal{J}_{\mathcal H}$ is the discrete-time penalty function corresponding to Eq.~\ref{eq:user define G absolute}.



\subsection{Inequality Constraints Elimination}
\label{Inequality Constraints Elimination}
Inspired by constraint transcription method \cite{jennings1990computational}, the inequality constraints Eq.~\ref{eq:user define G relative} and  Eq.~\ref{eq:user define G absolute} are formulated into penalty functions. 

\subsubsection{Continuous Relative-time Penalty}
For the constraint Eq.~\ref{eq:user define G relative} forced over the entire continuous trajectory, the derivatives between pieces are independent of each other.
We transform them into finite constraints via the integral of constraint violation, which is further transformed into the
penalized sampled function $\mathcal{J}_{\mathcal{I}}$:
\begin{subequations}
    \begin{align}
      &\mathcal{J}_{\mathcal{I}} =\int_{0}^{T}  \mathcal{J}_{\mathcal G} dt=\sum_{i=1}^M \mathcal{I}_i,\\
      &\mathcal{I}_i=\frac{T_i}{\kappa_i} \sum_{j=0}^{\kappa_i} \bar{\omega}_j  \mathcal{J}_{\mathcal{G}} \left(\mathbf z^{[s-1]}_i(t_i),\bm \varrho(t_i)\right),
    \end{align}
\end{subequations}
where ${\kappa_i}$ is the sample number, $\left(\bar{\omega}_0, \bar{\omega}_1, \ldots, \bar{\omega}_{\kappa_i-1}, \bar{\omega}_{\kappa_i}\right)=(1 / 2,1, \cdots, 1,1 / 2)$ are the quadrature coefficients following the trapezoidal rule \cite{Press2007numerical}, and $t_i=\frac{j}{\kappa_i} T_i$.
Then the gradient of $\mathcal{I}_i$ w.r.t. $\textbf{c}$ and $\textbf{T}$ can be easily derived:
\begin{subequations}
    \begin{align}
        \frac{\partial \mathcal{J}_{\mathcal{I}}}{\partial \textbf{c}_i}&= \frac{T_i}{\kappa_i} \sum_{j=0}^{\kappa_i} \left( \bar{\omega}_j  \frac{\partial \mathcal{J}_{\mathcal G}}{\partial \mathbf{c}_i} \right), \\
        \frac{\partial \mathcal{J}_{\mathcal{I}}}{\partial T_i}&=\frac{\mathcal{I}_i}{T_i}+ \frac{T_i}{\kappa_i}  \sum_{j=0}^{\kappa_i} \left( \bar{\omega}_j \frac{j}{\kappa_i}  \frac{\partial \mathcal{J}_{\mathcal G}}{\partial t_i} \right).
    \end{align}
\end{subequations}
For $\partial\mathcal{J}_{\mathcal G} / {\partial \textbf{c}_i}$, the total derivative of it is written as
\begin{equation}
\frac{\partial \mathcal{J}_{\mathcal G}}{\partial \textbf{c}_i} = \sum_{k=0}^{s-1} \bm \beta^{\left(k\right)} \frac{\partial \mathcal{J}_{\mathcal G}}{\partial \mathbf z_i^{\left(k\right)}}.
\end{equation}
Similar for $\partial\mathcal{J}_{\mathcal G} / {\partial t_i}$ that can be written as
\begin{equation}
\frac{\partial \mathcal{J}_{\mathcal G}}{\partial t_i} = \sum_{k=0}^{s-1} \mathbf z_i^{\left(k+1\right)} \frac{\partial \mathcal{J}_{\mathcal G}}{\partial \mathbf z_i^{\left(k\right)}} + \dot{\bm \varrho} \frac{\partial \mathcal{J}_{\mathcal G}}{\partial \bm \varrho} + \dot{\theta} \frac{\partial \mathcal{J}_{\mathcal G}}{\partial \theta}.
\end{equation}

\subsubsection{Discrete Absolute-time Penalty}
To ensure consistency with the discrete results of the path-finding method, we formulate the corresponding constraint Eq.~\ref{eq:user define G absolute} forced on the trajectory over specific discrete moments.
The cost of a piece of the trajectory at a discrete moment will be affected by the duration changes of the former pieces. 
For a state $\mathbf z(t_k)$, assumed that $t_k$ is an absolute moment on the j-th piece of the trajectory $\mathbf z_j(t)$, the relative time $t_r = (t_k-\sum_{i=1}^{j-1} \textbf{T}_i)$ on $\mathbf z_j(t)$ is within the range
\begin{equation}
\sum_{i=1}^{j-1} T_i \leq t_k \leq \sum_{i=1}^j T_i.
\end{equation}
With the denotation of $\mathcal{J}_{\mathcal D} = \sum_{t \in \mathcal T} \mathcal{J}_{\mathcal H}$ in Eq.~\ref{equ:unconstainted problem}, the gradients of $\mathcal{J}_{\mathcal D}$ w.r.t. $\textbf{c}$ and $\textbf{T}$ can be evaluated as
\begin{subequations}
\begin{align}
\label{eq:dHdc}
\frac{\partial \mathcal{J}_{\mathcal D}}{\partial \textbf{c}_i} & = \begin{cases}
\sum_{k=0}^{s-1}\bm \beta^{\left(k\right)} \left(t_r\right) {\partial \mathcal{J}_{\mathcal D}}/{\partial \mathbf z_i^{\left(k\right)}}, & i=j, \\
0, & i \neq j,\end{cases} \\
\label{eq:dHdT}
\frac{\partial \mathcal{J}_{\mathcal D}}{\partial T_i} & = \begin{cases}
\sum_{k=0}^{s-1}-\mathbf z_i^{\left(k+1\right)}\left(t_r\right) 
{\partial\mathcal{J}_{\mathcal D}}/{\partial \mathbf z_i^{\left(k\right)}}, & i<j, \\
0, & i \geq j.\end{cases}
\end{align}
\end{subequations}




Then, the problem of calculating gradients w.r.t $\textbf{c}$ and $\textbf{T}$ is further transformed into taking the gradients w.r.t the drone's flat-output states and their derivatives $\mathbf z^{\left[ s-1\right]}$.

\subsection{Temporal Constraints Elimination}
\label{Temporal Constraints Elimination}
Time slack constraint Eq.~\ref{eq:T>Tl} can be written as
\begin{equation}
\sum_{i=1}^M T_i \geq T_{\ell}.
\end{equation}
We denote $\boldsymbol{\varsigma}=\left(\varsigma_1, \ldots, \varsigma_M\right) \in \mathbb{R}^M$ as new variables to be optimized and use the transformation
\begin{subequations}
\begin{align}
T_{\Sigma} & =T_l+\varsigma_M^2 \\
T_i & =\frac{e^{\varsigma_i}}{1+\sum_{j=1}^{M-1} e^{\varsigma_j}} T_{\Sigma}, 1 \leq i<M \\
T_M & =T_{\Sigma}-\sum_{i=1}^{M-1} T_i.
\end{align}
\end{subequations}
The temporal constraints are eliminated by such substitution.
Especially, when $T_{\ell}=0$, we use a simpler transformation $T = e^{\varsigma}$ to eliminate the constrain Eq.~\ref{eq:T>Tl}. Since the zero duration is invalid, here constraint Eq.~\ref{eq:T>Tl} is changed to $T>T_{\ell}$.



\subsection{Tracking Constraints Formulation}
\label{subsec: tracking constraints}
For trajectory optimization for tracking, we adopt $\mathfrak{T}^3_{\mathrm{MINCO}}$ to represent position trajectory and $\mathfrak{T}^2_{\mathrm{MINCO}}$ to represent yaw trajectory, guaranteeing sufficient optimization freedom. 
The differentiable visibility metrics have been presented in Sec.~\ref{sec:Visibility-aware Tracking Metrics}, and the constraints for tracking are further clarified here. Since the front end offers discrete drone and target series, the corresponding visibility costs forced at specific times are discrete absolute-time penalties. Then the continuous relative-time and discrete absolute-time penalties are listed as
\begin{subequations}
    \begin{align}
    &\mathcal{J}_{\mathcal G} = \lambda_{\mathcal G} \left[ \mathcal J_{o}, \mathcal J_{v}, \mathcal J_{a}\right]^T,\\
    &\mathcal{J}_{\mathcal H} = \lambda_{\mathcal H} \left[ \mathcal J_{OD}, \mathcal J_{OA},\mathcal J_{OE} \right]^T,
    \end{align}
\end{subequations}
where $ \lambda_{\mathcal G}$ and $ \lambda_{\mathcal H}$ are preset weight vectors. $\mathcal J_{OD}$ is the observation distance penalty function defined as Eq.~\ref{equ:do_cost}, $\mathcal J_{OA}$ is the observation angle part defined as Eq.~\ref{equ:ao_cost}, and $\mathcal J_{OE}$ is the occlusion effect avoidance part defined as Eq.~\ref{equ:eo_cost}.
In addition, $\mathcal J_{o}$ is obstacle avoidance cost defined as
\begin{equation}
\label{equ:P_occ}
    \mathcal J_{o}(t) = g({d_{thr}^2} - \Xi(\textbf{p}(t))^2),
\end{equation}
where $d_{thr}$ is the safety threshold, and $\Xi(\cdot)$ is the distance to the closest obstacle obtained from ESDF.
$\mathcal J_{v}, \mathcal J_{a}$ are dynamic feasibility costs:
\begin{equation}
\label{equ:P_dv}
    \mathcal J_{v}(t) = g(\norm{\textbf{p}^{(1)}(t)}^2 - v_{max}^2),
\end{equation}
\begin{equation}
\label{equ:P_da}
    \mathcal J_{a}(t) = g(\norm{\textbf{p}^{(2)}(t)}^2 - a_{max}^2),
\end{equation}
where $v_{max}$ and $a_{max}$ are the maximum velocity and acceleration, respectively.
For temporal constraint Eq.~\ref{eq:T>Tl}, the duration of tracking trajectory should be equal to the prediction duration of the target motion $T_p$, ideally. However, enforcing the drone to reach the final states in a fixed duration may cause dynamic infeasibility, for instance, when the target moves faster than the drone. Therefore, we make
a time slack $T \ge T_p$.
For the final state constraint Eq.~\ref{eq:final pos}, we fix it to the last viewpoint $\mathbf{v}_K$. 
For trajectory initialization, the initial guess of $\textbf{m}$ is obtained by sampling on the path provided by the occlusion-aware path-finding method, and the initial $T_i=T_p/M$.

\begin{figure}[b]
    \begin{center}
         \includegraphics[width=0.95\columnwidth]{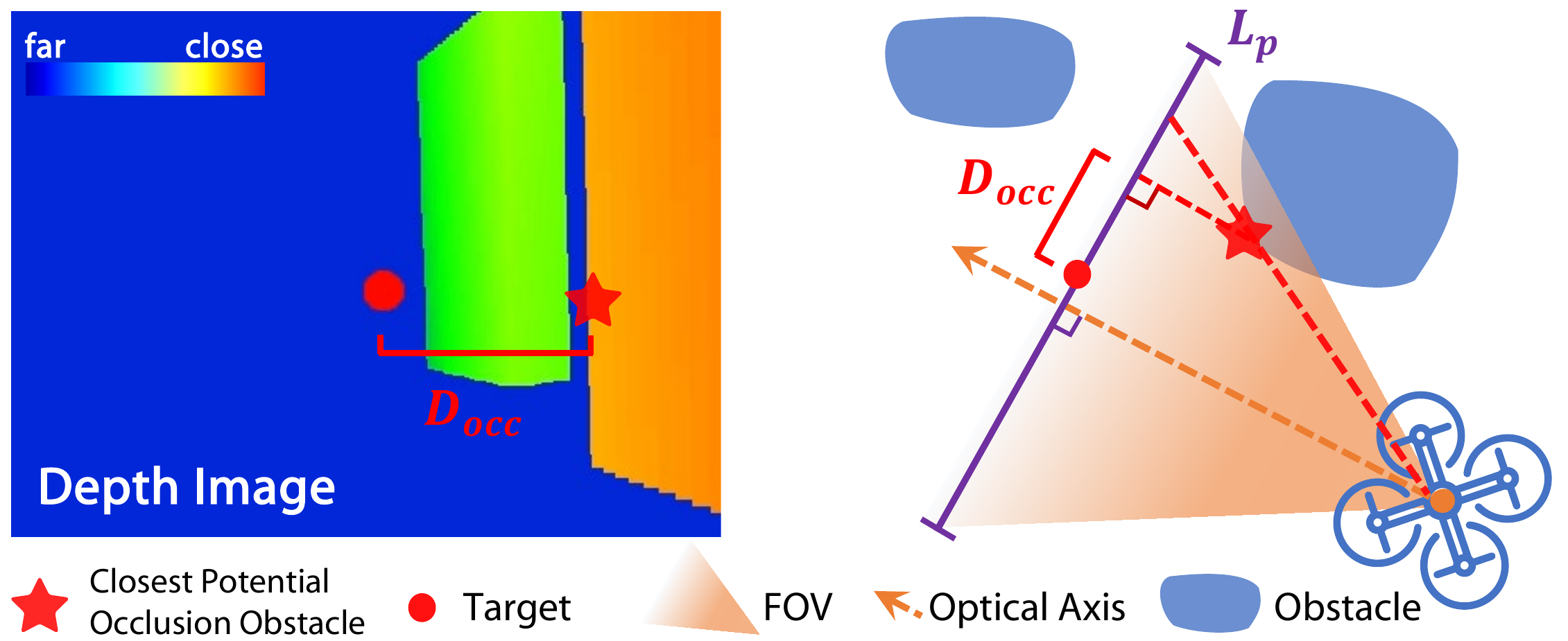}
    \end{center}
    \caption{Illustration of the occlusion criterion $D_{occ}$, which denotes the closest possible occluding obstacle. $L_p$ is the projected plane crossing the target. Only obstacles on the camera image between the drone and $L_p$ are recorded.}\label{fig:OE_metric}
\end{figure}

\subsection{Perching Constraints Formulation}
\label{subsec: perching constraints}
For trajectory optimization for perching, we adopt $\mathfrak{T}^4_{\mathrm{MINCO}}$ to represent position trajectory and $\mathfrak{T}^2_{\mathrm{MINCO}}$ to represent yaw trajectory. When perching, it is necessary to force the terminal angular velocity to be $0$, and $\mathfrak{T}^4_{\mathrm{MINCO}}$ is the lowest order MINCO supporting to constrain the terminal angular velocity~\cite{wang2022geometrically}. 
With initial time set as $T_i=\norm{\mathbf{p}-\boldsymbol{\varrho}}/v_{max}$, a boundary value problem is solved to obtain the initial guess of $\textbf{m}$.
Then, we summarize the penalty functions as
\begin{subequations}
    \begin{align}
    &\mathcal{J}_{\mathcal G} = \lambda_{\mathcal G} \left[\mathcal J_{c}, \mathcal J_{p}, \mathcal J_{\omega}, \mathcal J_{\tau}, \mathcal J_{\Delta z}, \mathcal J_{\nu} \right]^T,
    \end{align}
\end{subequations}
where $\mathcal J_{c}$ is collision avoidance cost defined in Eq.~\ref{SE3 cost function}, $\mathcal J_{p}$ is perception cost defined in Eq.~\ref{perception cost function}. $\mathcal J_{\omega}, \mathcal J_{\tau}$ are dynamic feasibility cost defined in Eq.~\ref{omega cost function} and Eq.~\ref{tau cost function}, $\mathcal J_{\Delta z}$ is relative height cost defined in Eq.~\ref{eq:relative z cost function}, $\mathcal J_{\nu}$ is the regulation for tangential relative speed defined in Eq.~\ref{eq:relative vt}. 
\begin{figure*}[t]
    \begin{center}
         \includegraphics[width=1.94\columnwidth]{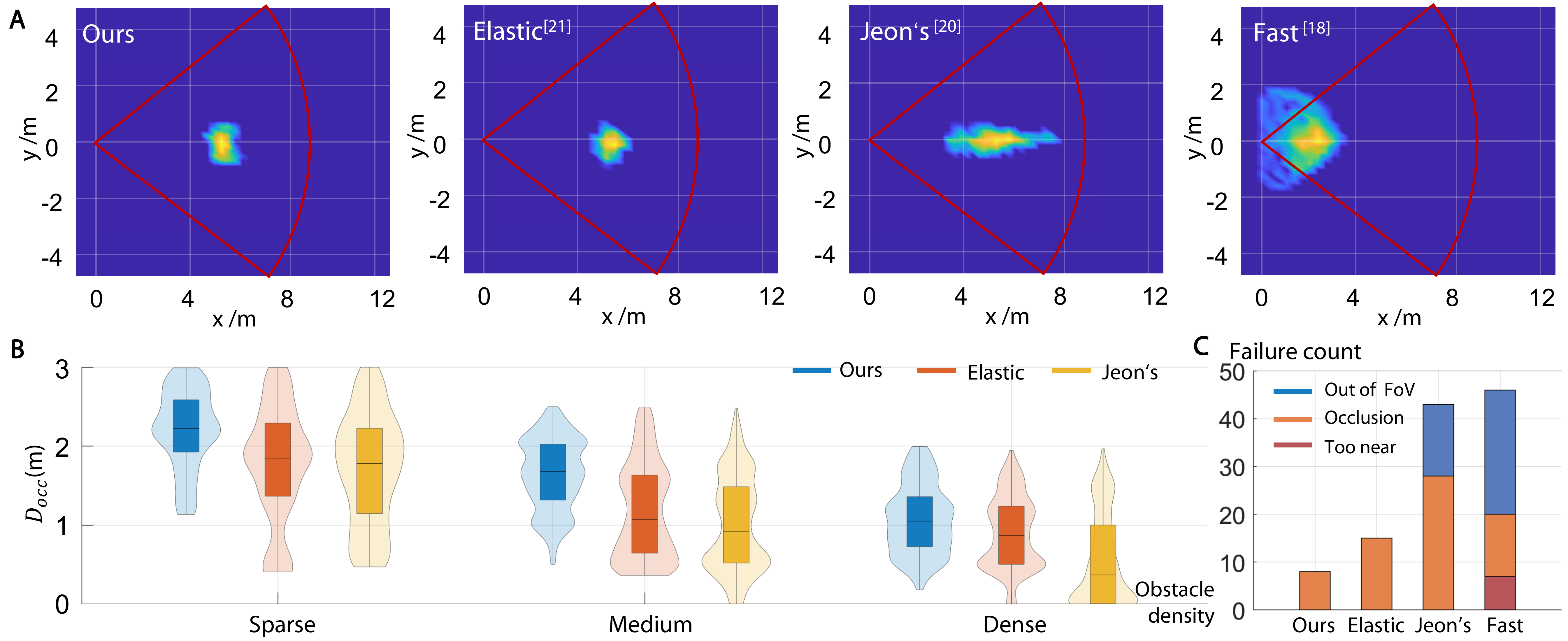}
    \end{center}
    \vspace{-0.3 cm}
    \caption{\textbf{A}. Comparison of the distribution heat map of the target position relative to the tracking quadrotor. The red sector represents the FoV of the drone. \textbf{B}: The violin graph presenting a comparison of the distribution of the closest possible occluding obstacle on the image plane ($D_{occ}$) under different obstacle densities. \textbf{C}: The failure reasons including out of view, too near (distance less than $1m$), occluded by obstacles.}\label{fig:general_test}
\end{figure*}
For temporal constraint Eq.~\ref{eq:T>Tl}, the low bound is set to $T_{\ell} = 0$.





\section{Simulation and Benchmark}
\label{sec:benchmark}

In this section, we conduct extensive simulations to validate the performance of our tracking and perching planning methods. As for benchmarks, cutting-edge works \cite{han2021fast, jeon2020integrated,ji2022elastic,paneque2022perception,mao2023robust} in the tracking and perching fields are selected. All the simulations run on an Intel Core i5-9400F CPU with GeForce GTX 2060 GPU.

\subsection{General Comparison of Visibility-aware Tracking}
\label{subsec:tracking_general}

We benchmark our method with Jeon's \cite{jeon2020integrated} work and Ji's Elastic-Tracker \cite{ji2022elastic}, which both consider the target's visibility. To further demonstrate the importance of visibility awareness, we also compare with Han's Fast-Tracker \cite{han2021fast} without visibility consideration. These works are all open source and have not been modified.
To compare the tracking performance fairly, we set the same parameters as listed in Tab.~\ref{tab:para setting} with the same target trajectory.
To compare the capability of occlusion avoidance, since the drone obtains target information directly from camera images, we record the projected distance between the target and the closest possible occluding obstacle ($D_{occ}$). The occlusion criterion is shown in Fig.~\ref{fig:OE_metric}.

\begin{table}[ht]
	\renewcommand\arraystretch{1.2}
	\centering
	\caption{Parameters Setting for Tracking Simulations}
	\label{tab:para setting}
        \setlength{\tabcolsep}{1.0mm}{
	\begin{tabular}{cccccc}
        \toprule
        $v_{max}$ & $a_{max}$ & $\norm{\omega_{max}}$  & FoV  &   \makecell[c]{Image\\resolution} & \makecell[c]{Tracking\\distance}\\
        \midrule
         \makecell[c]{Drone: $3.0 \, \text{m}$\\Target: $1.5 \, \text{m}$}  &\makecell[c]{Drone: $6.0 \, \text{m/s}^2$\\Target: $2.0 \, \text{m/s}^2$}  & $3.0 \, \text{rad/s}$      & \makecell[c]{$80^{\circ}\times$\\$65^{\circ}$} & \makecell[c]{$640\times$\\$480 \text{ px}^2$} & $5.5 \, \text{m}$  \\
        \bottomrule
	\end{tabular}}
\end{table}

We test all four methods in the same simulation environment as Elastic-Tracker. Since obstacle density has a significant impact on tracking performance, we test in scenarios with different obstacle densities. The sparse, medium and dense density correspond to $6 \, \text{m}$, $5 \, \text{m}$ and $4 \, \text{m}$ average obstacle spacing, respectively.
During tracking, we count the target positions projected to x-y plane in the tracking quadrotor's FoV. In Fig.~\ref{fig:general_test}A, the heat map shows the distribution of the target positions, relative to the tracking drone. Our method and Elastic-Tracker can both keep the target in the proper position of the FoV. Fig.~\ref{fig:general_test}B shows the distribution of $D_{occ}$. Since the main failures for Fast-Tracker are caused by tracking too close to the target or letting the target out of FoV, making the $D_{occ}$ criterion meaningless, we do not include it in Fig.~\ref{fig:general_test}B.
Our method results in larger $D_{occ}$ in all obstacle densities and is more effective at avoiding potential occlusion.

Furthermore, we record the failure reasons during tracking. Once the target is out of FoV, too near to the target, or occluded by obstacles, a failure is recorded. The failure reasons in the dense obstacle scene are counted in Fig.~\ref{fig:general_test}C.
\begin{figure*}[t]
    \begin{center}
         \includegraphics[width=1.8\columnwidth]{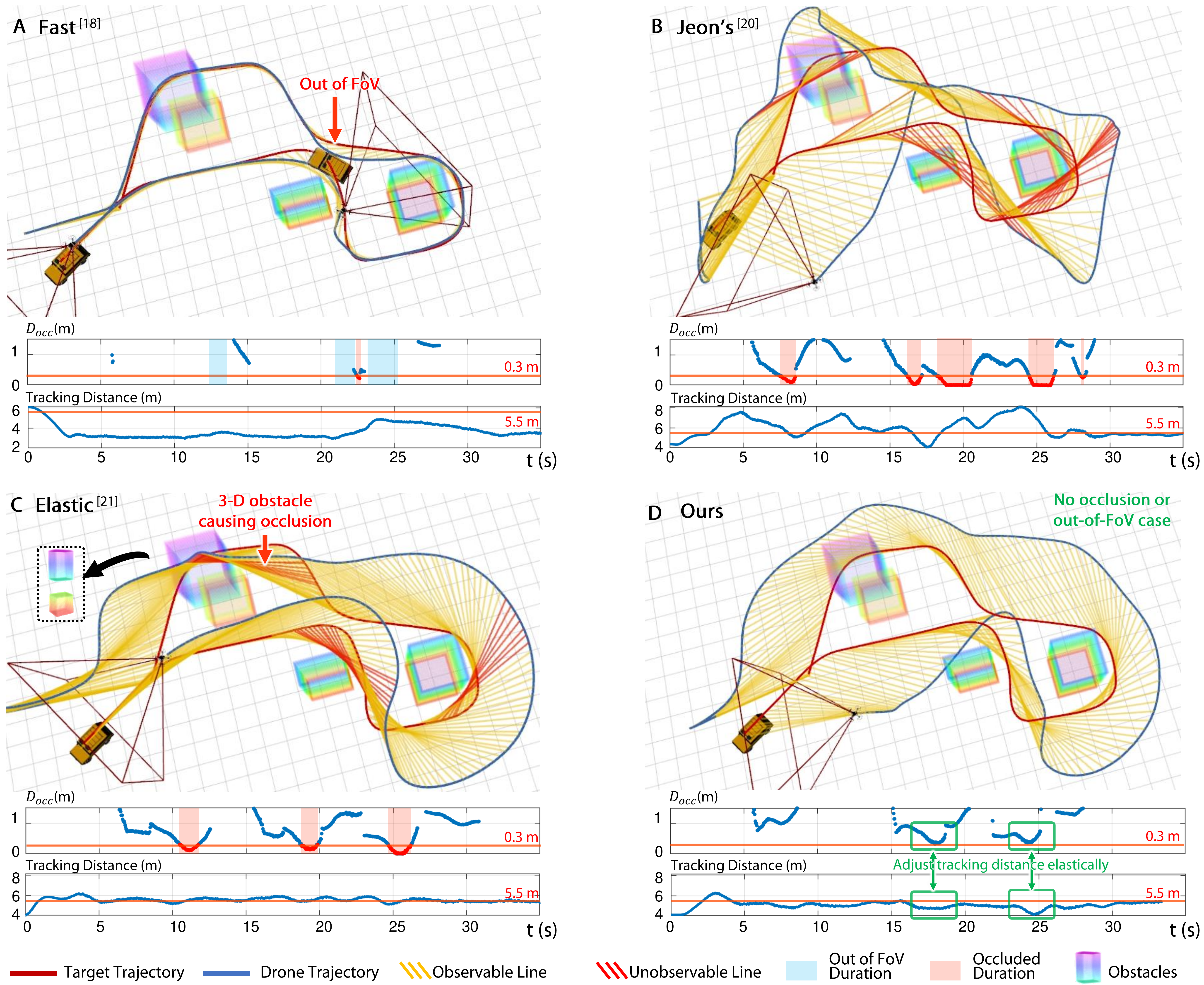}
    \end{center}
    \vspace{-0.3 cm}
    \caption{Case study of visibility-aware tracking. Trajectories of the drone and the target, and the lines connecting their positions at the corresponding time are presented. The red lines show the occlusion moments. The $D_{occ}$ and the tracking distance are shown for all four methods. We choose $0.3 m$ as the threshold at which the occlusion is likely to occur. An out-of-FoV failure of Fast-Tracker is shown in Fig.~A. Fig.~C shows that Elastic-Tracker fails to adjust visibility against the 3-D obstacle due to its pseudo-3-D visibility metric. Fig.~D shows that our method has no occlusion and out-of-FoV situations, and can adjust the relative distance elastically according to the environment.}\label{fig:case_study}
    
\end{figure*}
We also benchmark the computation time, including path finding and trajectory optimization parts, as shown in Tab.~\ref{tab:visibility_time}. Fast-Tracker consumes less time for trajectory optimization since it ignores visibility, resulting in more failures. In comparison, our proposed method consumes a much lower total computation budget, due to our lightweight path-finding method and concise metrics for trajectory optimization. 
\begin{table}[ht]
	\renewcommand\arraystretch{1.2}
	\centering
	\caption{Computation time Comparison between Tracking Methods}
	\label{tab:visibility_time}
        \setlength{\tabcolsep}{4.9mm}{
	\begin{tabular}{cccc}
        \toprule
        \multicolumn{1}{c}{\multirow{2}{*}{Methods}} & \multicolumn{3}{c}{Average time (ms)} \\
        \cmidrule(lr){2-4} 
        \multicolumn{1}{c}{}                           & $t_{path}$ & $t_{opt}$  & $t_{total}$         \\
        \midrule
        \multicolumn{1}{c}{Fast}                & 9.56           & 0.38            & 9.94      \\
        \multicolumn{1}{c}{Jeon's}            & 11.32            & 3.20            & 14.32       \\
            \multicolumn{1}{c}{Elastic}              & 1.24           & 2.34            & 3.58       \\
            \multicolumn{1}{c}{Ours}              & 0.96           & 1.85            & \textbf{2.81}\\
        \bottomrule
	\end{tabular}}
\end{table}

\subsection{Case Study of Visibility-aware Tracking}
\label{subsec:case study}

To better demonstrate and compare the tracking performance of different methods, we set up a representative scenario and present the whole process of tracking in Fig.~\ref{fig:case_study}. The parameter setting is the same as Sec. \ref{subsec:tracking_general}. 
$D_{occ}$ and the tracking distance are shown for all four methods.
The occluded moments are indicated by the red lines connecting the drone and the target, and also correspond to the red background in the $D_{occ}$ curve.
For Fast-Tracker, due to the lack of visibility consideration, the quadrotor consistently follows the target closely. In the event of a sudden target turn, the drone often fails to adjust quickly, resulting in the target escaping from the FoV, as the moment shown in Fig.~\ref{fig:case_study}A.
As shown in Fig.~\ref{fig:case_study}B, Jeon's method results in multiple occlusion moments and less smooth trajectory, since the sequential graph-search-based path planner and smooth planner fail to jointly optimize visibility and smoothness and lack temporal optimization.
As shown in Fig.~\ref{fig:case_study}C, because Elastic-Tracker handles visibility by generating 2-D visible fans at a certain height, it fails to adjust visibility against the 3-D obstacle, which in this case is a two-layer instance. 
More importantly, when passing obstacle-rich areas, our method can adjust tracking distance elastically to reduce occlusion probability, as highlighted in green in Fig.~\ref{fig:case_study}D.
Especially when tracking the target passing the narrow gap, the drone with our method deployed elastically reduce the tracking distance to $4 \, \text{m}$ and avoid occlusion successfully.
Consequently, our method shows reasonable visibility adjustment and has better performance at avoiding occlusion, due to the comprehensive and effective 3-D visibility metrics.

\begin{figure}[t]
    \begin{center}
         \includegraphics[width=0.97\columnwidth]{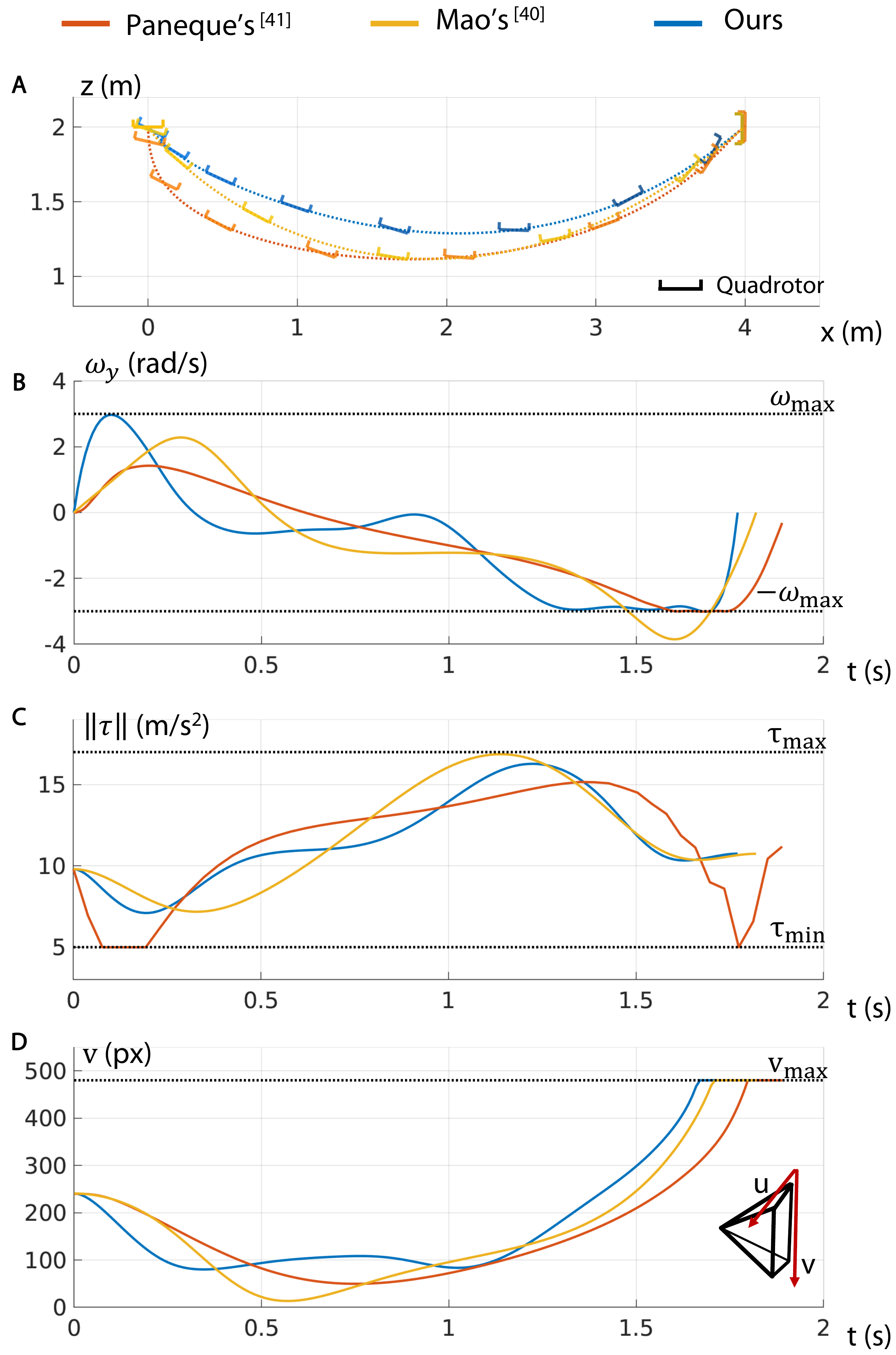}
    \end{center}
    \caption{Comparison of the trajectories generated by Paneuqe’s, Mao's, and our planner for perching on vertical surfaces. \textbf{A}. The x-z trajectory shape. \textbf{B}. The angular velocity. Mao's method exceeds the limitation. \textbf{C}. The mass-normalized net thrust. Due to the finite discrete resolution of Paneuqe’s method, its thrusts are less smooth than the others. \textbf{D}. The target position on the u-v image space. All the methods guarantee that the target is visible.}\label{fig:perch_com}
\end{figure}

\begin{figure}[t]
    \begin{center}
         \includegraphics[width=0.93\columnwidth]{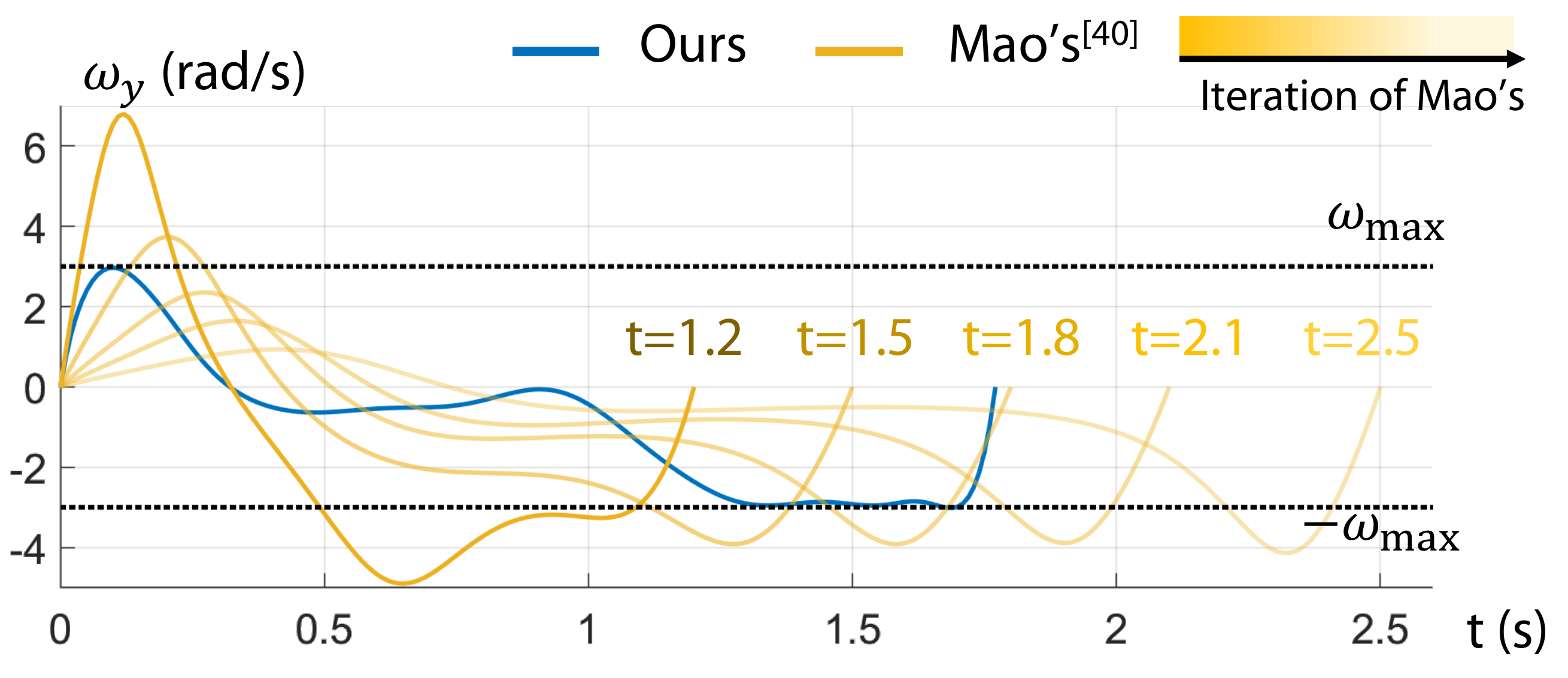}
    \end{center}
    \caption{Comparison of the angular velocity of the trajectories generated by our method and Mao's method. Mao's method fails to guarantee dynamic feasibility even after sufficient iterations to increase trajectory duration.
    }\label{fig:mao_fail}
\end{figure}

\subsection{Perching Benchmark}
\label{subsec: perching benchmark}
We benchmark our perching method with the state-of-the-art perching methods proposed by Paneque \cite{paneque2022perception} and Mao \cite{mao2023robust}, which both consider visibility during perching. 
Paneque formulates a constrained discrete-time NLP problem and solves it using ForcesPRO \cite{FORCESPro}. 
They model the inputs of the system as the desired constant thrust derivatives and control each rotor thrust directly. 
To compare fairly, we constrain the thrust as the same as us so that the dynamics model of Paneque's and ours are similar. 
Since this work focuses on the perception of powerlines, we replace its perception cost with ours to compare fairly.
Mao formulates a QP problem to constrain the terminal state and velocity bound. They increase the trajectories’ time iteratively and recursively solve the QP problem until the thrust constraint is satisfied.
We use the open-source code of Paneque‘s work and reimplement Mao's work which is not open-source but not difficult to implement with the QP formulation.
In the comparison, the terminal pitch angle is set to $\pi/2 \, \text{rad}$, and the trajectories are required to be rest-to-rest. 
The parameters of the drone are listed in Tab.~\ref{tab:para setting perching}.

\begin{table}[ht]
	\renewcommand\arraystretch{1.2}
	\centering
	\caption{Parameters Setting for Perching Simulations}
	\label{tab:para setting perching}
        \setlength{\tabcolsep}{2.4mm}{
	\begin{tabular}{ccccc}
        \toprule
        $\tau_{min}$ & $\tau_{max}$ & $\norm{\omega_{max}}$  & FoV  &   \makecell[c]{Image resolution}\\
        \midrule
         $5.0 \, \text{m/s}^2$  &$17.0 \, \text{m/s}^2$  & $3.0 \, \text{rad/s}$      & \makecell[c]{$80^{\circ}\times 65^{\circ}$} & \makecell[c]{$640\times 480 \text{ px}^2$}  \\
        \bottomrule
	\end{tabular}}
\end{table}

\begin{table}[ht]
	\renewcommand\arraystretch{1.2}
	\centering
	\caption{Computation Time Comparison between Perching Methods}
	\label{tab:perch_time}
        \setlength{\tabcolsep}{4.9mm}{
	\begin{tabular}{cccc}
        \toprule
        \multicolumn{1}{c}{\multirow{2}{*}{Methods}} & \multicolumn{3}{c}{Average time ($ms$)} \\
        \cmidrule(lr){2-4} 
        \multicolumn{1}{c}{}                           & $30^\circ$ & $60^\circ$  & $90^\circ$         \\
        \midrule
        \multicolumn{1}{c}{Paneque's}            & 239.59           & 308.05            & 208.81       \\
            \multicolumn{1}{c}{Mao’s (A)}              & 3.42           & 2.36            & \XSolidBrush       \\
            \multicolumn{1}{c}{Mao’s (B)}              & 16.65           & 13.73            & \XSolidBrush       \\
            \multicolumn{1}{c}{Ours}              & 3.76           & 4.31            & 16.68\\
        \bottomrule
	\end{tabular}}
\end{table}

\begin{figure}[b]
    \begin{center}
         \includegraphics[width=0.9\columnwidth]{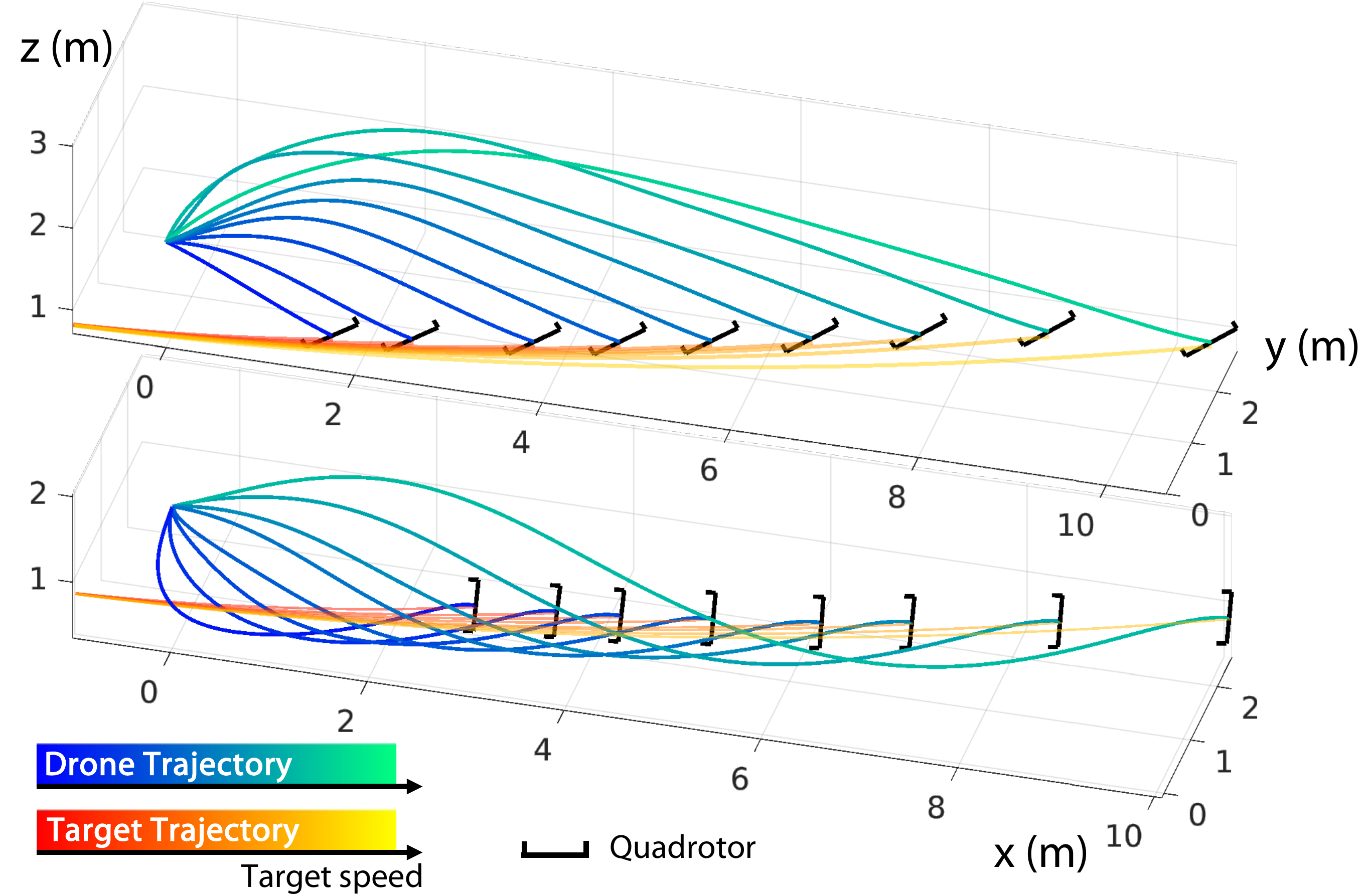}
    \end{center}
    \caption{Illustration of the three-dimensional trajectories generated for perching on a moving plane with different speeds. \textbf{Top}: $0.5 \, \text{rad}$ terminal pitch angle. \textbf{Bottom}: $1.5 \, \text{rad}$ terminal pitch angle. The target's velocity is ranged from $1.5 \, \text{m/s}$ to $3.0 \, \text{m/s}$. The kinematics of the target conforms to the CTRV model with $\omega= 0.2\, \text{rad/s}$.
    }\label{fig:perch_3d}
\end{figure}

We compare both the trajectory quality and computation time of these methods, the results are presented in Fig.~\ref{fig:perch_com} and Tab.~\ref{tab:perch_time}. Paneque's and our method successfully generate perching trajectories satisfying actuator and visibility constraints. Nevertheless, Paneque's method consumes much more computation time than ours, since a complex discrete-time multiple-shooting NLP problem formulation. 
Adopting MINCO, our method can optimize trajectories in the low-dimension flat-output space with high efficiency and can achieve real-time replanning during agile flight. 
Moreover, due to the finite discrete resolution, its thrusts and body rates are less smooth than ours. 
As shown in Fig.~\ref{fig:mao_fail}, Mao's method fails to guarantee dynamic feasibility even after sufficient iterations, which double the duration without providing any relief. Mao's method ignores that dynamic feasibility is affected by not only the temporal profile but also the spatial profile. As a result, its strategy narrows the solution space, leading to no solutions under such tight actuator constraints.
Moreover, Mao's computation time depends on both initial duration and time sampling resolution. Thus, for Mao's (A), we set a better initial duration $1.5 \, \text{s}$ and a coarser resolution $0.1 \, \text{s}$, while for Mao's (B), a worse initial duration $1.0 \, \text{s}$ and a finer resolution $0.05 \, \text{s}$ are set. The results indicate that the computation time of Mao's method greatly increases with a worse initial value and a finer resolution. 
In comparison, our proposed method demonstrates high efficiency without greatly sacrificing the optimality of solutions.

\begin{figure*}[ht]
    \begin{center}
         \includegraphics[width=1.9\columnwidth]{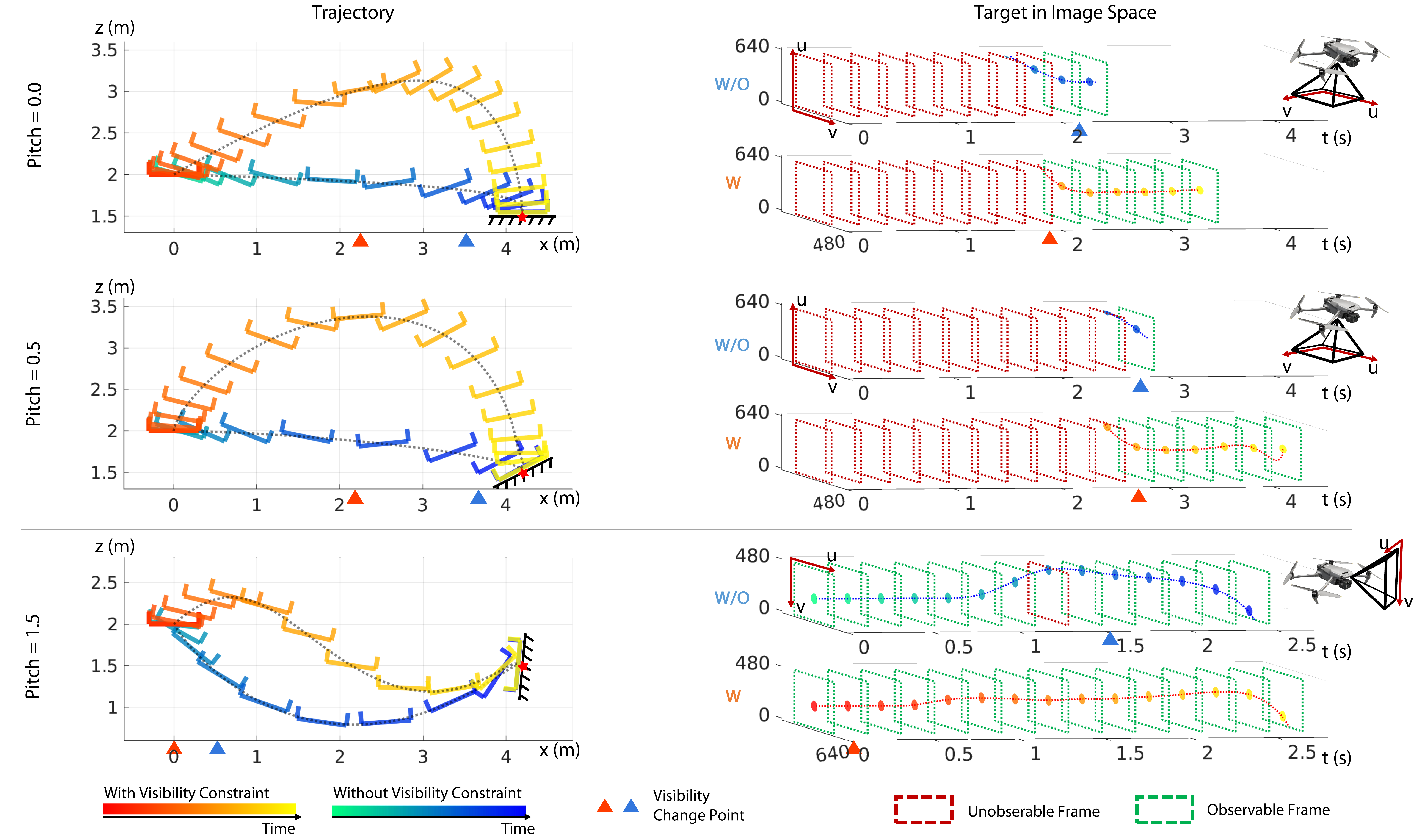}
    \end{center}
    \vspace{-0.3 cm}
    \caption{Ablation study of the visibility awareness at different perching terminal angles. The quadrotor uses the down camera to observe the target at $0.0$ and $0.5 \, \text{rad}$ terminal pitch angles, and the front camera for $1.5\, \text{rad}$ terminal pitch angles. The expected observe distance is set to $\bar{d}_{min} = 0.2\, \text{m}$ and $\bar{d}_{max} = 2.0\, \text{m}$. \textbf{Left}: The x-z trajectory w/ and w/o perception constraint. The triangles on the x-axis represent the position when the spatial visibility changes. The target is visible on the right side of the point. \textbf{Right}: The target position in the u-v image space. The target position is expected to be as close to the image's center as possible. The triangles on the x-axis represent the moment when the temporal visibility changes.}\label{fig:ablation study}
\end{figure*}

Furthermore, we compare the capability of our perching planner with Paneuqe’s
planner, Mao's planner, and our previous planner \cite{ji2022real}.
The result is summarized in Table \ref{tab:perch_capability}.
Both Paneuqe’s and Mao's methods can only handle static targets. 
While ours can simultaneously handle collision avoidance, dynamic feasibility, and visibility constraints, and generate perching trajectories toward a moving target at a low computation budget. 
To further demonstrate our capability for adapting dynamic targets, Fig.~\ref{fig:perch_3d} presents the generated trajectories for perching on moving targets with different speeds. 
Balanced aggressiveness and dynamic feasibility, our method drives the quadrotor perch on the target at proper contact moments.
With different contact moments and different terminal states, the quadrotor could still get full-state synchronization with the perching surface, presenting flexibility.


\begin{table}[ht]
    \caption{Capability Comparison between Perching Methods}
    \label{tab:perch_capability}
    \centering
    \begin{tabular}{ccccc}

    \toprule
    \diagbox[width=7em]{Method}{Capability} & \makecell[c]{Collision\\Avoidance}   & \makecell[c]{Dynamic\\Feasibility} & \makecell[c]{Visibility\\Awareness} & \makecell[c]{Dynamic\\Target}  \\
    \toprule
     Paneque's & \Checkmark & \Checkmark & \Checkmark & \XSolidBrush   \\
     Mao's & \XSolidBrush &\Checkmark & \Checkmark & \XSolidBrush  \\
     Ji's & \Checkmark &\Checkmark & \XSolidBrush & \Checkmark  \\
     \textbf{Ours} & \Checkmark & \Checkmark & \Checkmark & \Checkmark  \\
   \bottomrule
    \end{tabular} 
\end{table}

\subsection{Ablation Study of the Visibility Awareness for Perching}
\label{subsec:ablation study}

\begin{figure}[ht]
	\begin{center}
		\includegraphics[width=1.0\columnwidth]{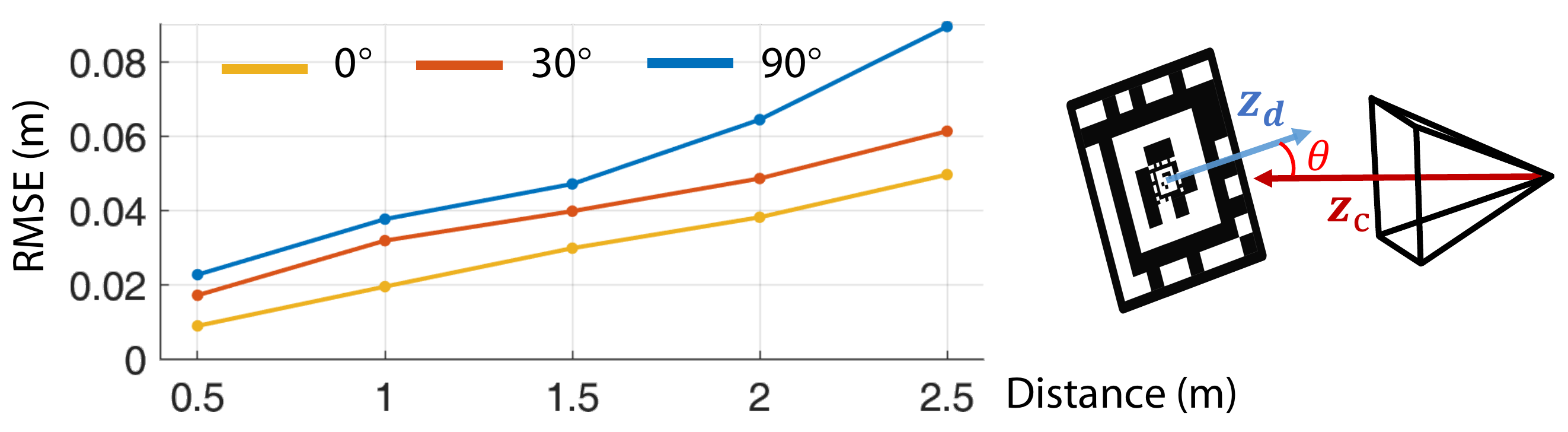}
	\end{center}
	\caption{
	\label{fig:tag_err}
	Visual tag localization error at different distances and observation angles $\theta$. $\mathbf{z}_d$ is the normal vector of the tag plane and $\mathbf{z}_c$ is the optical axis.
	}
\end{figure}

Continuous and stable perception of targets is the foundation for perching planning.
We validate our visibility-aware approach by comparing the perching trajectories with and without the visibility term.
We use AprilTag \cite{krogius2019flexible} to mark the precise perching position. As shown in Fig.~\ref{fig:tag_err}, localization error increases with the observation angle $\theta$.
The quadrotor we used has front and down cameras for observation. Therefore, we choose the down camera for continuous detection when the terminal attitude is below $30^{\circ}$ and the front camera when it is above $30^{\circ}$. Moreover, the decreasing error with decreasing distance also presents the necessity of replanning. 
\begin{figure*}[ht]
    \begin{center}
        \includegraphics[width=1.88\columnwidth]{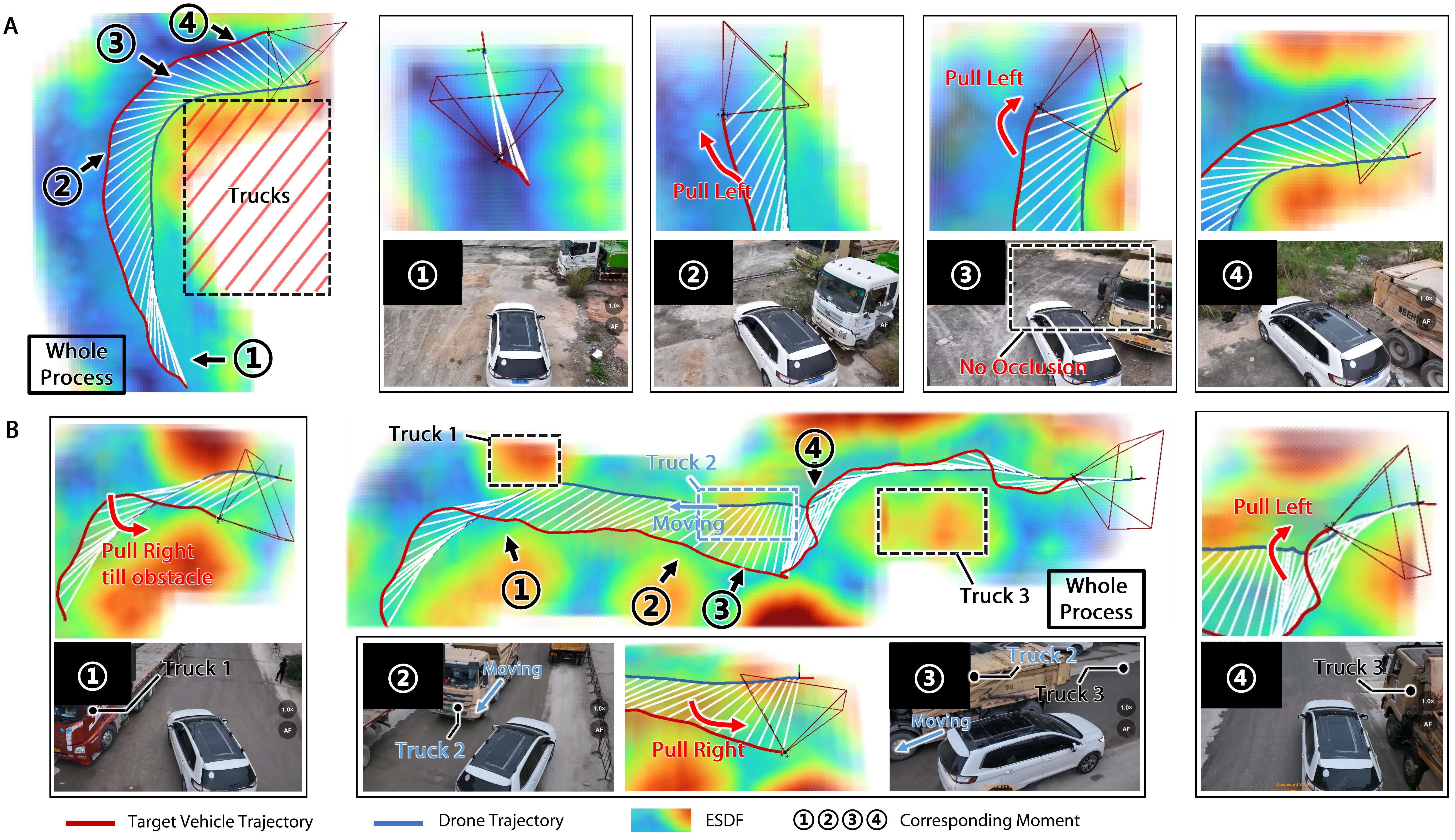}
    \end{center}
    \vspace{-0.3 cm}
    \caption{Real-world visibility-aware tracking validation. \textbf{A. L-shape Turn}: \ding{172} The drone tracks the SUV behind. \ding{173} a trunk appears closely on the right. The drone pulls left to avoid occlusion. \ding{174} The SUV turns right along the trunk, easily causing occlusion. The drone pulls left to avoid it. \ding{175} Constantly tracking on the left for better visibility. \textbf{B. Bilateral Staggered Obstacles}: \ding{172} The SUV heading straight, truck 1 appears on the left,  and the drone pulls right. \ding{173} Trunk 2 is approaching the target SUV from the left. \ding{174} The SUV stops to make way for truck 2. The tracking drone keeps pulling right as truck 2 gets closer. \ding{175} The SUV passes truck 3, and the drone pulls left for less occlusion.}
    \label{fig:vis_exp_1}
\end{figure*}

We conduct tests under different terminal attitudes, as shown in Fig.~\ref{fig:ablation study}. 
The parameters are the same as Sec.~\ref{subsec: perching benchmark}. 
The expected observation range is set to $\bar{d}_{min} = 0.2\, \text{m}$ and $\bar{d}_{max} = 2.0\, \text{m}$.
The orange and blue trajectories show quadrotor maneuvers with and without visibility considerations, respectively. 
For $0.0 \, \text{rad}$ and $0.5 \, \text{rad}$ terminal pitch angles, without visibility consideration, the drone can only observe the target at a short distance and a short duration. While with the perception cost, the quadrotor can keep continuous observation within the preset observe distance $\bar{d}_{max} = 2.0\, \text{m}$. For $1.5\, \text{rad}$ pitch angle, the visibility-aware perching can keep the target at the center of the image space as long as possible. The observable distance and duration before touching the target plane are summarized in Tab.~\ref{tab:visibility_perch}. 
Consequently, with our visibility-aware perching planning, the target visibility in terms of both space and time is dramatically improved.

\begin{table}[ht]
    \caption{Ablation Study of the Visibility Awareness for Perching}
    \label{tab:visibility_perch}
    \centering
    \begin{tabular}{  c  c  c  c }
    \toprule
    \makecell[c]{Terminal\\Pitch Angle} & \makecell[c]{Visibility\\Awareness}& \makecell[c]{Observable\\Distance (m)}& \makecell[c]{Observable\\Duration (s)}\\
    \toprule
    \multirow{2}{*}{\textbf{0.0 rad}}
    &\multirow{2}{*}{\makecell[c]{w/o\\w/}}  
    &\multirow{2}{*}{\makecell[c]{0.64\\ \textbf{1.96}} \! ($\uparrow$203\%)} 
    &\multirow{2}{*}{\makecell[c]{0.60\\ \textbf{1.51}} \! ($\uparrow$152\%)}\\
    \\
    \midrule

    \multirow{2}{*}{\textbf{0.5 rad}}
    &\multirow{2}{*}{\makecell[c]{w/o\\w/}}  
    &\multirow{2}{*}{\makecell[c]{0.44\\ \textbf{1.99}} \! ($\uparrow$352\%)} 
    &\multirow{2}{*}{\makecell[c]{0.36\\ \textbf{1.68}} \! ($\uparrow$367\%)}\\
    \\
    \midrule

    \multirow{2}{*}{\textbf{1.5 rad}}
    &\multirow{2}{*}{\makecell[c]{w/o\\w/}}  
    &\multirow{2}{*}{\makecell[c]{3.61\\ \textbf{4.15}} \! ($\uparrow$15\%)} 
    &\multirow{2}{*}{\makecell[c]{2.2\\ \textbf{2.96}} \! ($\uparrow$35\%)}\\
    \\
   \bottomrule
    \end{tabular} 
\end{table}


\section{Real world experiments}
\label{sec:experiments}
\subsection{System Configuration}
\begin{figure}[b]
    \begin{center}
        \includegraphics[width=0.85\columnwidth]{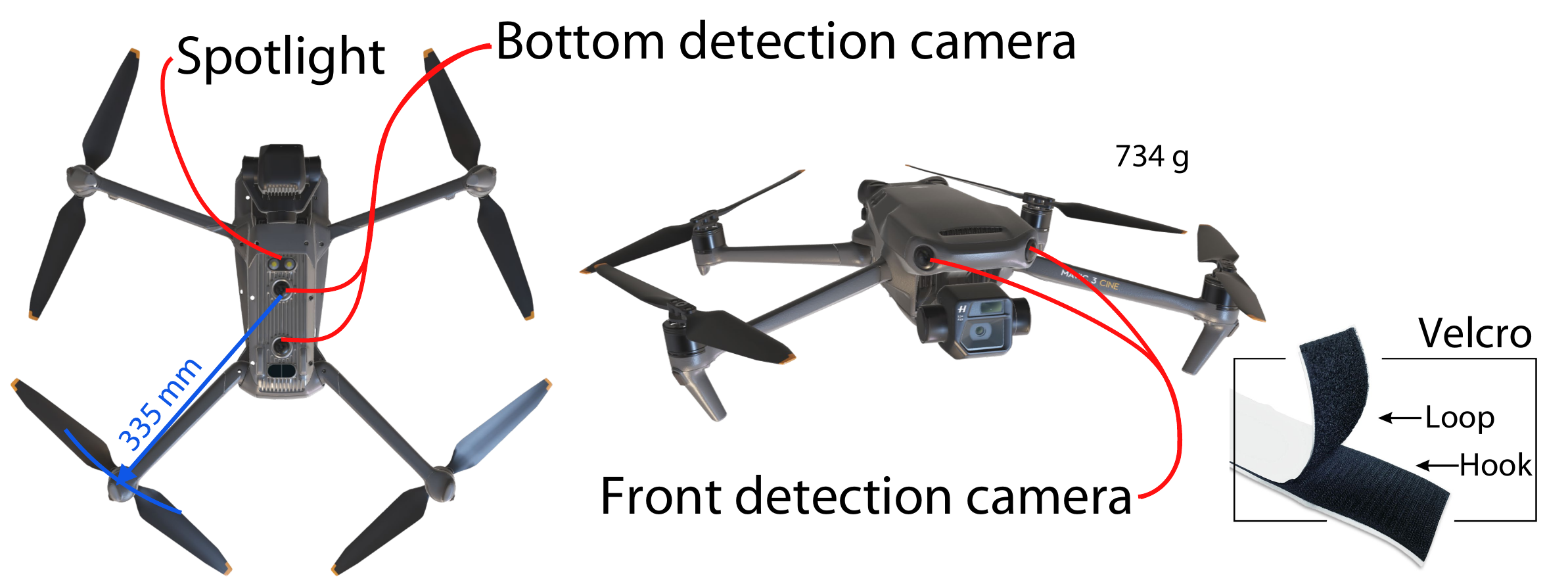}
    \end{center}
    \caption{Illustration of the quadrotor platform (DJI MAVIC3). The front and bottom cameras are used for detection. The gimbal camera was not selected for detection since its dedicated designed for photography and with high link delay. The drone and the surface are secured with velcro after perching.}
    \label{fig:drone}
\end{figure}

We deploy our adaptive tracking and perching scheme on a commercial drone (DJI MAVIC3) shown in Fig.~\ref{fig:drone}, with a full-size SUV as the moving platform. 
Our system operates outside the laboratory and is tested on the road, bringing it closer to real-world applications.
The drone will attempt to track and perch on the SUV. The perching stage starts when receiving external perching order.

The target position is obtained by the fusion of finer visual detection and coarser non-station differential-GPS-based relative localization. We use the recursive visual fiducial tags \cite{krogius2019flexible} to mark the precise perching position. 
The software implementation of our scheme is integrated into the drone's embedded processors, and interfaces with the existing modules to form a complete system. The localization and mapping functionality of the DJI MAVIC3 is retained as the foundation of the system. After the generation of the trajectory, control instructions are derived and transmitted to the flight controller for execution. Software modules, estimation, perception, planning, and control are all running onboard in real time. On the severe-limited computing platform ($8 \times$ slower than simulation settings), our planning scheme can still achieve $20\, \text{Hz}$ replanning.

\subsection{Visibility-aware Tracking Validation}
We select two common real-world scenarios to validate the visibility awareness of our tracking scheme.
Static and moving impediments will appear in the path of the vehicle when it moves, stops, or turns.
The drone is expected to track the vehicle without occlusion or target loss. 


\subsubsection{Scenario A: L-shape Turn}
The occlusion frequently happens when the target makes turns around obstacles.
In this case, the target vehicle makes an L-shape turn, and two trucks are parked at the bend, which could easily block the view of the tracking drone. As shown in Fig.~\ref{fig:vis_exp_1}A, due to our visibility-aware tracking scheme, the drone actively adjusts its relative position and keeps its FoV free of occlusion even when the vehicle makes sudden turns near obstacles.

\subsubsection{Scenario B: Bilateral Staggered Obstacles}
While the SUV is moving, static or dynamic trucks will appear alternately on the left and right sides. The target vehicle will either stop or move forward. In this experiment, not only the active adjustment of the drone to the static surroundings is demonstrated, but also the ability to avoid dynamic obstacles affecting the observation line of sight.
As shown in Fig.~\ref{fig:vis_exp_1}B, the tracking drone maintains high visibility of the target SUV throughout the process.

\subsubsection{Comparative Experiment}
\begin{figure}[ht]
    \begin{center}
         \includegraphics[width=1.0\columnwidth]{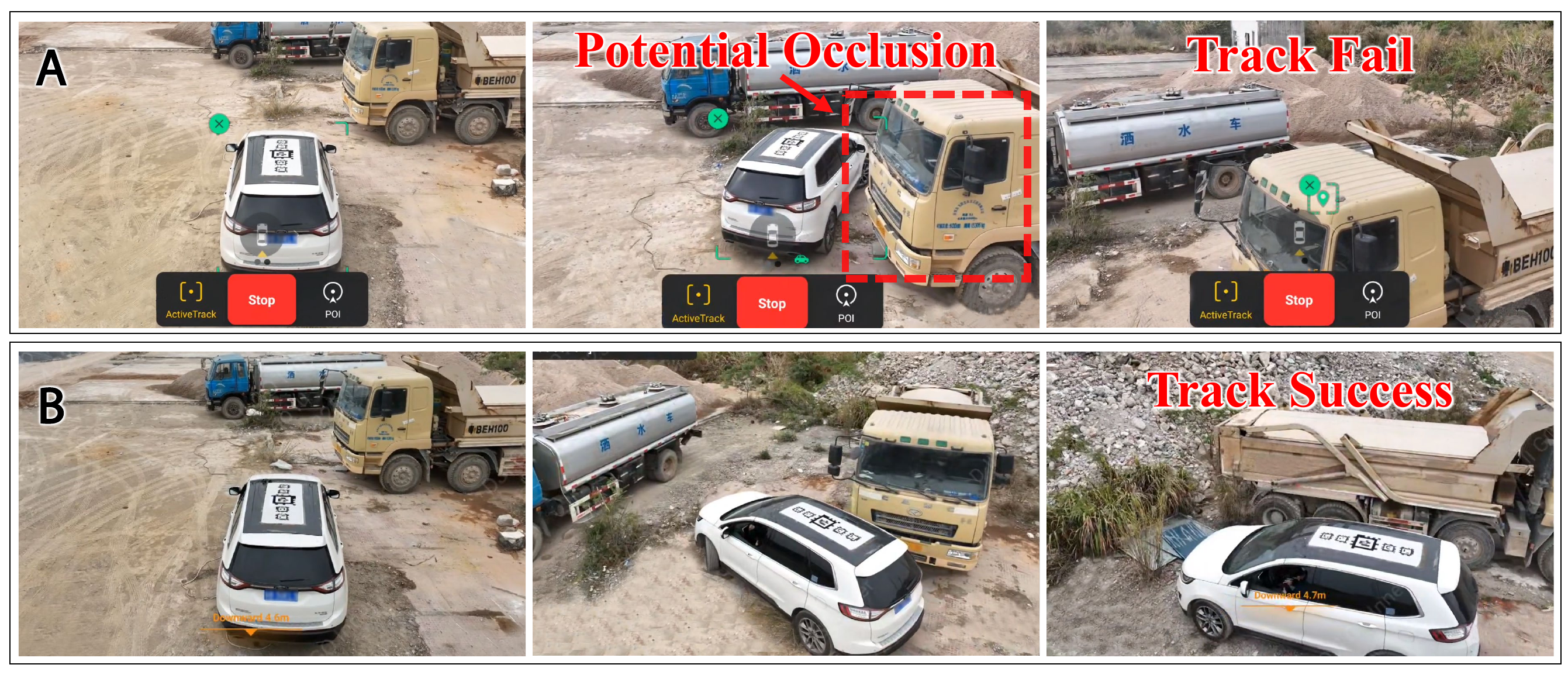}
    \end{center}
    \caption{Comparison with the origin DJI tracking under Scenario 2. \textbf{A}. The \textit{Active Tracking} function of DJI MAVIC loses the target car at the turning. \textbf{B}. Our tracking scheme enhances visibility actively, tracking successfully.}\label{fig:L_compare} 
\end{figure}

We compare our tracking scheme with the mature commercial \textit{Active Tracking} function of DJI MAVIC3 under Scenario A, as shown in Fig.~\ref{fig:L_compare}. When the target SUV turns right near the truck, the drone with \textit{Active Tracking} is unaware of the potential occlusion and loses the target at the turning. While our method drives the drone to avoid the occlusion effect of the truck, and allows the drone to continuously receive visual information of the target.

In these experiments, the drone observes the target SUV at a certain distance with a straight-forward angle and effectively avoids the occlusion of obstacles. Consequently, our scheme drives the drone to maintain a stable relative state and high-quality observation with the target.

\begin{figure*}[ht]
    \begin{center}
        \includegraphics[width=1.85\columnwidth]{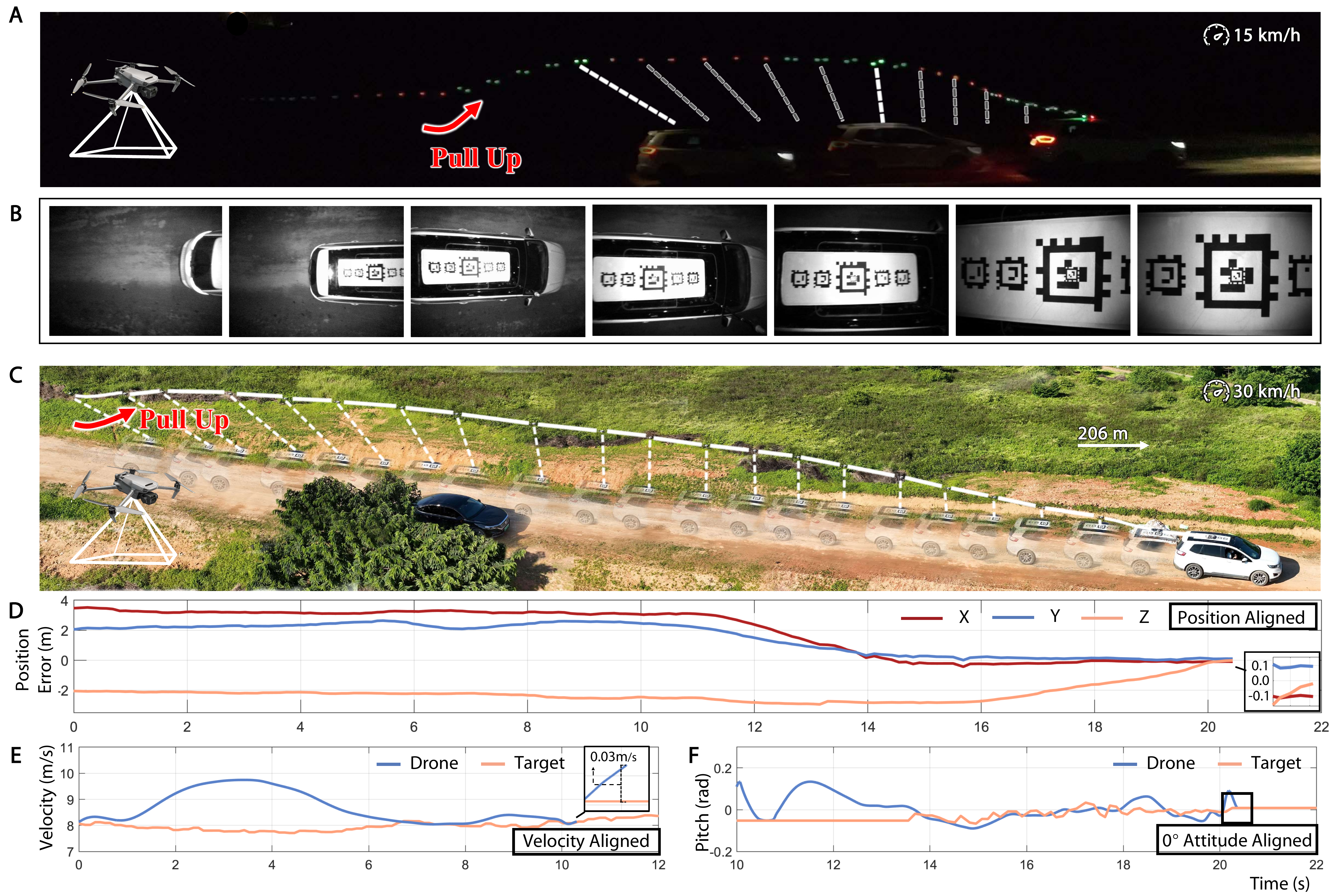}
    \end{center}
    \vspace{-0.5 cm}
    \caption{\textbf{A-B}.Illustration of the $15 \, \text{km/h}$ ($4.2 \, \text{m/s}$) tracking and horizontal perching real-world experiment in the night scene. \textbf{B}. The images obtained by the down camera of the drone. \textbf{C-F}.Illustration of the $30 \, \text{km/s}$ ($8.3 \, \text{m/s}$) high-speed tracking and horizontal perching. \textbf{D}. The position error between the drone and the desired perching point. \textbf{E}. The velocity curve of the drone and the SUV. \textbf{F}. The pitch angle curve of the drone and the perching surface on the SUV.}
    \label{fig:landing1}
    \vspace{-0.3 cm}
\end{figure*}


\subsection{High-speed Dynamic Tracking and Perching Validation}
Extensive experiments are conducted to validate dynamic tracking and perching in multiple scenarios. The system is tested in high-speed situations that require an efficient, robust, and accurate tracking and perching scheme. The following tests present the system's high precision, system-level robustness, and active visibility enhancement. 
We first test horizontal perching in multiple scenarios.
The results of the experiments are summarized in Tab.~\ref{tab:Multiple Scenes}. Indoor and road scenarios have good road conditions, yet the field road provides for safe high-speed tests but with more uneven ground causing target shaking. These abundant scenes also demonstrate the application value.
And the multiple desire terminal attitudes with different target speeds are also tested and summarized in Tab.~\ref{tab:Multiple Attitude}.
For both tables, the maximum and average velocity of the drone during perching are recorded. The average target acceleration estimation shows both real velocity changes and perception disturbance. Moreover, the final perching errors are listed in the table. We present three cases for more details as follows.

\begin{table}[t]
    \caption{Multiple Scenes Real-world Experiments}
    \label{tab:Multiple Scenes}
    \centering
    \begin{tabular}{  c  c  c  c  c }
    \toprule
    \makecell[c]{Scene} & \makecell[c]{Target Vel.\\(m/s)}& \makecell[c]{Drone Vel. (m/s)\\(Max. / Avg.)}& \makecell[c]{Target\\Acc. (m/s$^2$)} & \makecell[c]{Error\\(cm)}\\
    \toprule
    \multirow{1}{*}{\textbf{Indoor}}
    &\multirow{1}{*}{\makecell[c]{1.5}}  
    &\multirow{1}{*}{\makecell[c]{2.65 / 2.21} } 
    &\multirow{1}{*}{\makecell[c]{0.65} } 
    &\multirow{1}{*}{\makecell[c]{3.31} }
    \\
    \midrule

    \multirow{1}{*}{\textbf{Road}}
    &\multirow{1}{*}{\makecell[c]{4.0}}  
    &\multirow{1}{*}{\makecell[c]{5.13 / 4.76} } 
    &\multirow{1}{*}{\makecell[c]{1.01} } 
    &\multirow{1}{*}{\makecell[c]{4.47} }
    \\
    \midrule

    \multirow{1}{*}{\textbf{Field Night}}
    &\multirow{1}{*}{\makecell[c]{4.0 }}  
    &\multirow{1}{*}{\makecell[c]{5.22 / 4.47} } 
    &\multirow{1}{*}{\makecell[c]{1.67} } 
    &\multirow{1}{*}{\makecell[c]{8.36} }
    \\
    \midrule
    
    \multirow{3}{*}{\textbf{Field}}
    &\multirow{3}{*}{\makecell[c]{4.0\\ 6.0\\8.0}}  
    &\multirow{3}{*}{\makecell[c]{5.37 / 4.69\\ 7.53 / 6.87 \\ 9.51 / 8.44} } 
    &\multirow{3}{*}{\makecell[c]{1.12\\1.18\\ 1.31} } 
    &\multirow{3}{*}{\makecell[c]{5.87\\10.92\\ 14.19 \\ } }\\
    \\
    \\
    
   \bottomrule
    \end{tabular}
    
    

\end{table}

\begin{table}[t]
    \caption{Multiple Attitude Dynamic Perching Experiments}
    \label{tab:Multiple Attitude}
    \centering
    \begin{tabular}{  c  c  c  c  c }
    \toprule
    \makecell[c]{Pitch\\Angle} & \makecell[c]{Target Vel.\\(m/s)}& \makecell[c]{Drone Vel. (m/s)\\(Max. / Avg.)}& \makecell[c]{Target\\Acc. (m/s$^2$)} & \makecell[c]{Error\\(cm)}\\
    \toprule
    \multirow{3}{*}{\textbf{0.5 rad}}
    &\multirow{3}{*}{\makecell[c]{0.0\\ 1.0\\2.0}}  
    &\multirow{3}{*}{\makecell[c]{1.63 / 0.82\\ 1.73 / 1.51 \\ 3.04 / 2.33} } 
    &\multirow{3}{*}{\makecell[c]{0.27\\0.49\\ 0.60} } 
    &\multirow{3}{*}{\makecell[c]{2.05\\4.85\\ 6.29 \\ } }\\
    \\
    \\
    \midrule
    
    \multirow{3}{*}{\textbf{1.0 rad}}
    &\multirow{3}{*}{\makecell[c]{0.0\\ 1.0\\2.0}}  
    &\multirow{3}{*}{\makecell[c]{2.37 / 1.43\\ 2.84 / 1.91 \\ 3.87 / 2.86} } 
    &\multirow{3}{*}{\makecell[c]{0.26\\0.42\\ 0.41} } 
    &\multirow{3}{*}{\makecell[c]{2.86\\7.01\\ 9.33 \\ } }\\
    \\
    \\
    
   \bottomrule
    \end{tabular}

\end{table}

\subsubsection{Case 1: 15 km/h Tracking and Perching at Night}
In this case, we present the visibility-aware behavior resulting from our approach.
Thanks to the spotlight on the bottom of the drone, the bottom detection camera is adaptable to night scenes. The visual perception range is set to $[0.0\, \text{m},2.0\, \text{m}]$.

As shown in Fig.~\ref{fig:landing1}A, the initial relative height is less than $2.0\, \text{m}$, so the drone pulls itself up to ensure the observation of the target within the set range. Such behavior is not intentionally designed but rather emerges automatically due to the perception metric. The target tag is kept in the center of the image space as much as possible since its initial detection, which can enhance the target state estimation.

\subsubsection{Case 2: 30 km/h High-speed Tracking and Perching }
In this case, we conduct high-speed tracking and perching on the top of the SUV. During the perching process, as shown in Fig.~\ref{fig:landing1}B, the SUV's speed is $30\, \text{km/h}$ ($8.3 \, \text{m/s}$), while the maximum speed of the drone reaches about $10 \, \text{m/s}$. The final perching error on the $x-y$ plane is $10.7 \, \text{cm}$. The inaccuracy is mostly caused by poor target estimating (due to detection and system link delays) and trajectory tracking errors. The wind and the ground effect are countered by the integrator control. 
Due to the turbulence of the vehicle, the attitude of the perching surface is shaking. The drone can obtain the plane's attitude during continuous observation and constantly replan, and finally ensure the attitude alignment at the end. 

\begin{figure*}[ht]
    \begin{center}
         \includegraphics[width=1.85\columnwidth]{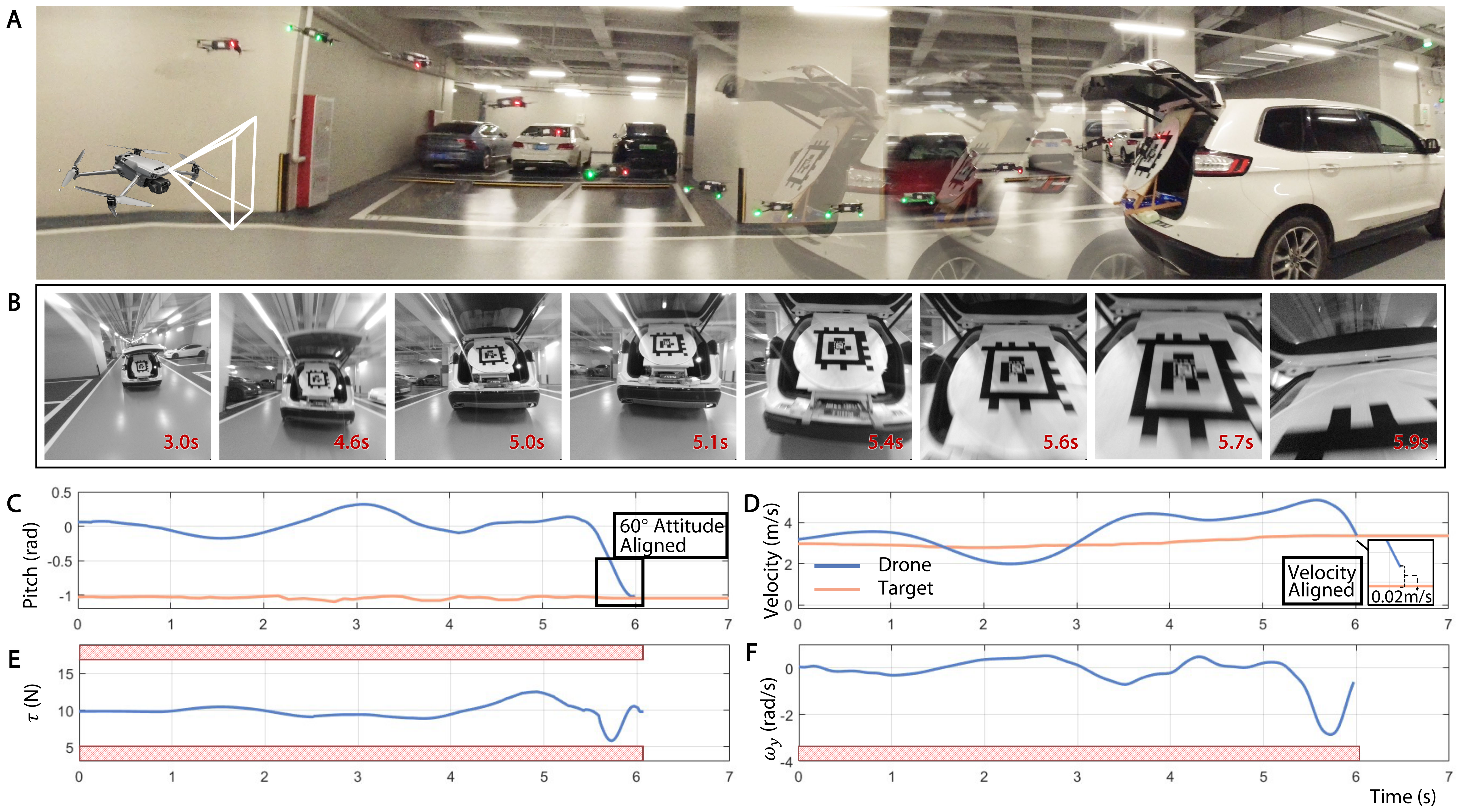}
    \end{center}
    \vspace{-0.5 cm}
    \caption{Illustration of the large attitude dynamic tracking and perching real-world experiment. \textbf{A}. The snapshot of the whole process. The vehicle moves at approximately $3.5\, \text{m/s}$, and the perching surface is $60^\circ$ inclined. \textbf{B}. The images obtained by the front camera of the quadrotor. \textbf{C}. The pitch angle of the drone and the moving plane. \textbf{D}. The velocity curve of the drone and the moving plane. \textbf{E}. The thrust curve of the drone. \textbf{F}. The angular velocity curve. }\label{fig:perching60}
    \vspace{-0.2 cm}
\end{figure*}

\subsubsection{Case 3: Large Attitude Dynamic Tracking and Perching}
In this case, we set a $60^{\circ}$ inclined plane in the trunk of the SUV as the perching surface to validate the large attitude perching ability. Fig.~\ref{fig:perching60} shows the snapshot of the whole process. For the sake of perception enhancement, the trajectory shape is similar to the simulated one with $1.5\, \text{rad}$ pitch angle shown in Fig.~\ref{fig:ablation study}. During the time interval from $3.0\, \text{s}$ to $5.7\, \text{s}$, the drone keeps the tag in the center of the image space as much as possible. The angular velocity and the thrust are both within the dynamic feasible region. 



\section{Conclusion}
\label{sec:conclusion}
In this paper, we analyze the core dilemmas to achieve tracking and perching in the dynamic scenario in detail and summarize five aspects of requirement accurately to solve the above problems.
Then we respectively design metrics to enhance visibility, guarantee safety, and dynamic feasibility in both tracking and perching stages. To adapt to the dynamic target state, we adjust the terminal state of the drone accordingly to achieve state alignment. High-frequency spatiotemporal SE(3) trajectory optimization provides feasible tracking and perching trajectories.
Finally, we deployed the adaptive tracking and perching scheme systematically on the commercial drone with a full-size SUV as the moving platform for real-world experiments. The results verify that our scheme is effective and has great application value.

\section{Acknowledgement}
The authors would like to thank colleagues at DJI for their strong support for this work: Zezao Lu, Zhongyan Xu, Bo Wu, Jialing Nong and Jiahang Ying.


\bibliography{main}

\vspace{-1.5cm}
\begin{IEEEbiography}[{\includegraphics[width=1in,height=1.25in,clip,keepaspectratio]{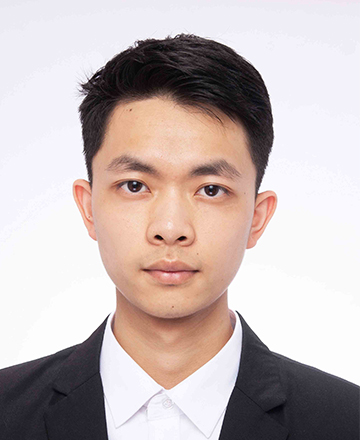}}]{Yuman Gao}
graduated from Zhejiang University, in 2021, at Zhejiang University, majoring in control science and engineering and concurrently pursuing a minor in ACEE (Advanced Honor Class of Engineering Education) at Chu Kochen Honors College. He's currently pursuing a Ph.D. in Automation under the supervision of Fei Gao, at the FAST Lab from Zhejiang University, China.
His research interests include motion planning and autonomous navigation for multi-robot systems.
\end{IEEEbiography}

\vspace{-1.2cm}
\begin{IEEEbiography}[{\includegraphics[width=1in,height=1.25in,clip,keepaspectratio]{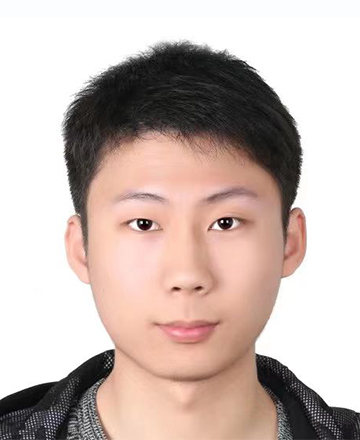}}]{Jialin Ji}
graduated from Zhejiang University, in 2019, with a major in mechatronic engineering at School of Mechanical Engineering and a minor in ACEE (Advanced Honor Class of Engineering Education) at Chu Kochen Honors College. He's currently pursuing a Ph.D. in Automation under the supervision of Fei Gao, at the FAST Lab from Zhejiang University, China, working on motion planning and autonomous navigation for aerial robotics.
\end{IEEEbiography}

\vspace{-1.2cm}
\begin{IEEEbiography}[{\includegraphics[width=1in,height=1.25in,clip,keepaspectratio]{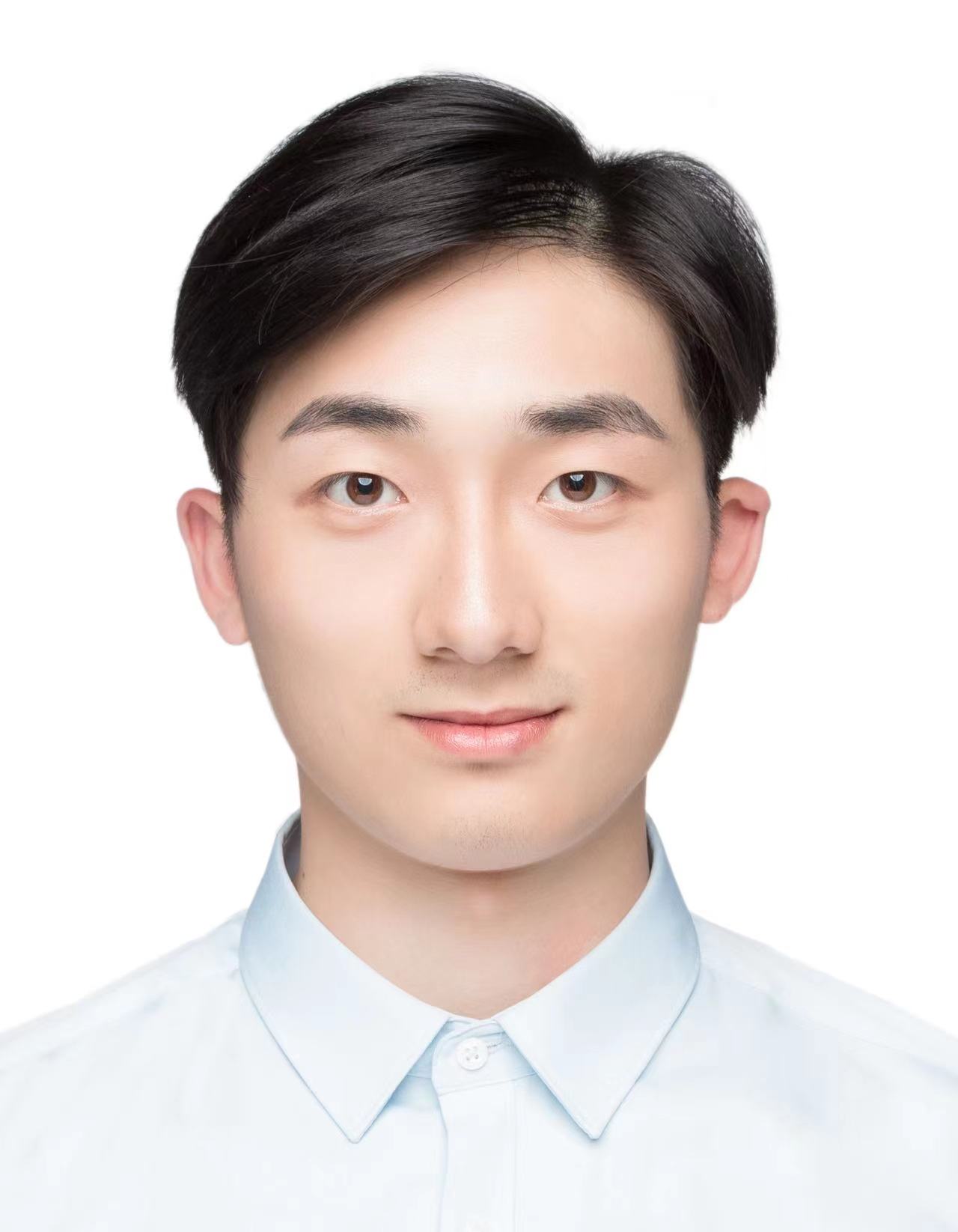}}]{Qianhao Wang}
received the M.Eng. degree in control science and engineering from Zhejiang University, Hangzhou, China, in 2022.  
He is currently working toward a Ph.D. degree in control science and engineering from Zhejiang University, Hangzhou, China. His research interests include motion planning, computational geometry, LiDAR SLAM, and autonomous navigation for aerial robotics.
\end{IEEEbiography}

\vspace{-1.2cm}
\begin{IEEEbiography}[{\includegraphics[width=1in,height=1.25in,clip,keepaspectratio]{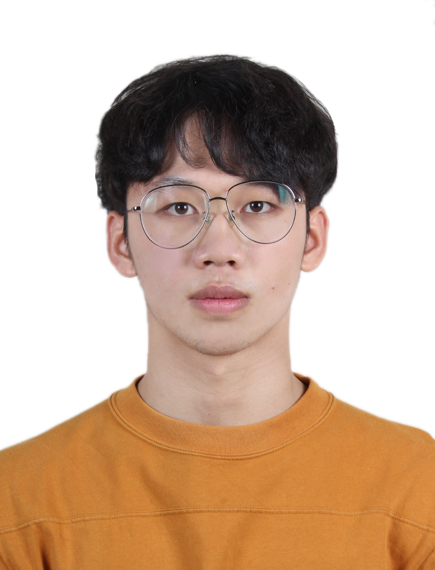}}]{Rui Jin}
received the B. E. degree from Northwestern Polytechnical University, in 2021 and is currently working toward the M. Eng. degree in robotics with the FAST Lab, Zhejiang University, Hangzhou, China. Her current research interests focus on the design, control, and motion planning of coaxial helicopters.
\end{IEEEbiography}

\vspace{-0.7cm}
\begin{IEEEbiography}[{\includegraphics[width=1in,height=1.25in,clip,keepaspectratio]{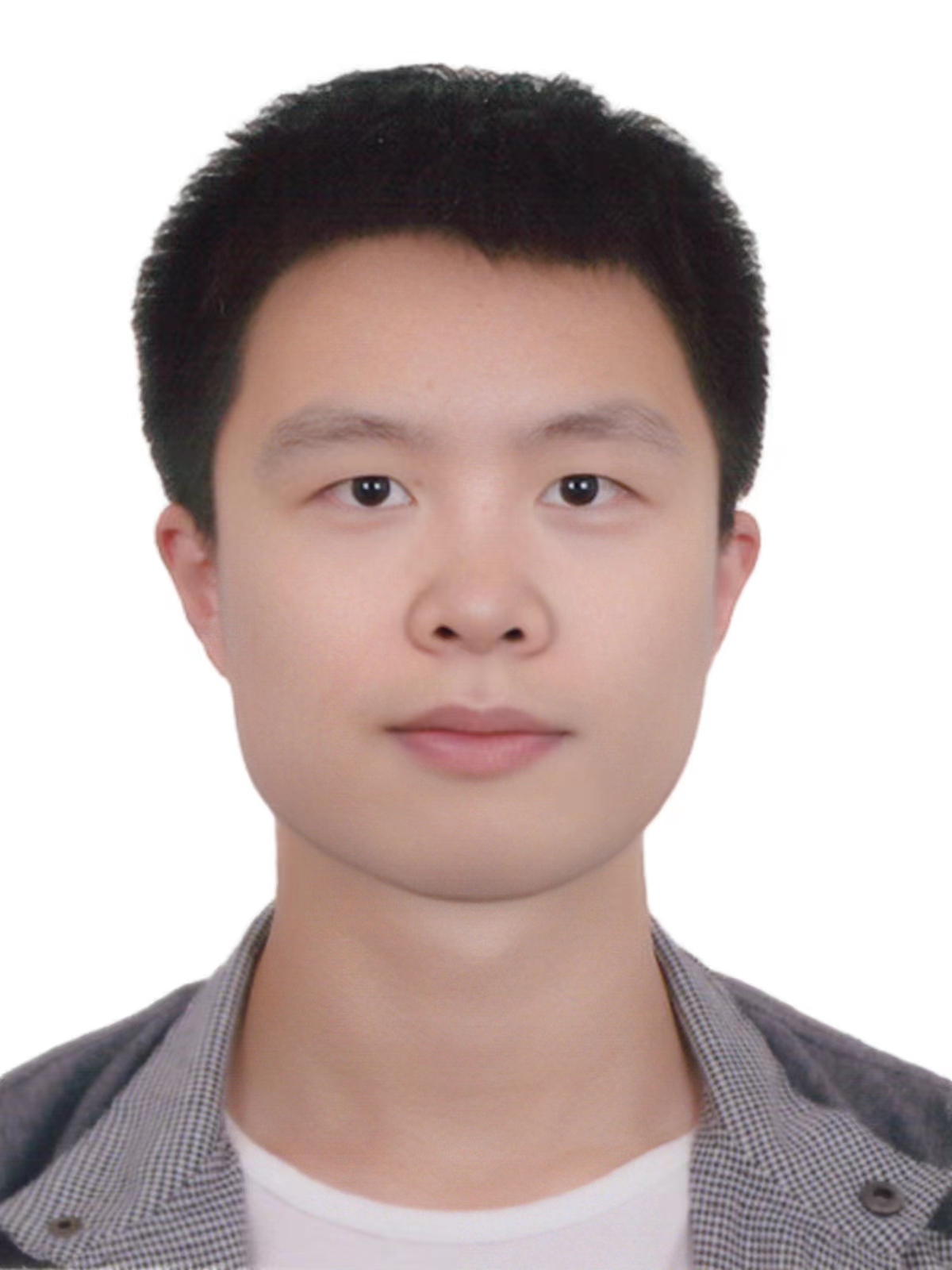}}]{Yi Lin}
received the B.Eng. degree in software engineering from Harbin Institute of Technology, Harbin, China, in 2015, and the M.Phil. degree in electronic and computer engineering from the Hong Kong University of Science and Technology, Hong Kong, in 2017.
He is currently a senior engineer at DJI Technology Co., Ltd. His research interests include computer vision, simultaneous localization and mapping, 3D reconstruction, autonomous navigation, unmanned aerial vehicles, motion planning, and optimization.
\end{IEEEbiography}

\vspace{-0.7cm}
\begin{IEEEbiography}[{\includegraphics[width=1in,height=1.25in,clip,keepaspectratio]{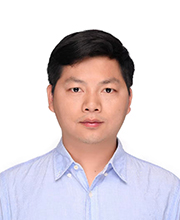}}]{Zhimeng Shang}
received the master degree in mechanical engineering and automation from Beihang University, Beijing, China in 2015.
He is currently working at DJI (since 2015) and focus on multirotor control and planning. He participated in the development of DJI's Mavic/Phantom/Inspire/FPV series, as well as flight controllers like A3/N3. His research interests include classic and optimal control, motion planning and uav design and control.
\end{IEEEbiography}

\vspace{-0.7cm}
\begin{IEEEbiography}[{\includegraphics[width=1in,height=1.25in,clip,keepaspectratio]{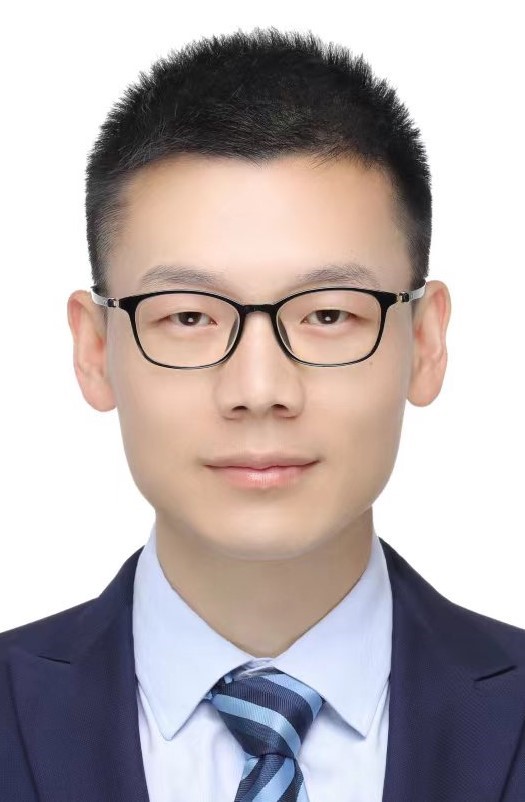}}]{Yanjun Cao}
received his Ph.D. in computer and software engineering from the University of Montreal, Canada, in 2020.
He is currently an associate researcher at the Huzhou Institute of Zhejiang University in the Center of Swarm Navigation. He leads the Field Intelligent Robotics Engineering (FIRE) group of the Field Autonomous System and Computing Lab (FAST Lab). 
His research focuses on key challenges in multi-robot systems, such as collaborative localization, autonomous navigation, perception, and communication.
\end{IEEEbiography}

\vspace{-0.7cm}
\begin{IEEEbiography}[{\includegraphics[width=1in,height=1.25in,clip,keepaspectratio]{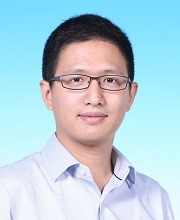}}]{Shaojie Shen}
received his B.Eng. degree in Electronic Engineering from the Hong Kong University of Science and Technology in 2009. He received his M.S. in Robotics and Ph.D. in Electrical and Systems Engineering in 2011 and 2014, respectively, all from the University of Pennsylvania, PA, USA. 
He joined the Department of Electronic and Computer Engineering at the HKUST in September 2014 as an Assistant Professor, and was promoted to Associate Professor in July 2020.
His research interests are in the areas of robotics and unmanned aerial vehicles, with focus on state estimation, sensor fusion, computer vision, localization and mapping, and autonomous navigation in complex environments.
\end{IEEEbiography}

\vspace{-0.7cm}
\begin{IEEEbiography}[{\includegraphics[width=1in,height=1.25in,clip,keepaspectratio]{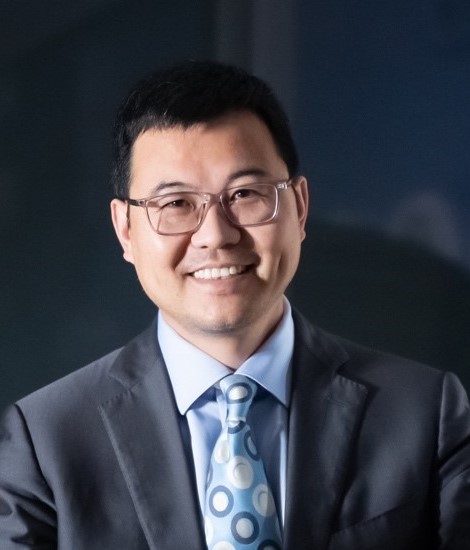}}]{Chao Xu}
received his Ph.D. in Mechanical Engineering from Lehigh University in 2010. He is currently Associate Dean and Professor at the College of Control Science and Engineering, Zhejiang University. He serves as the inaugural Dean of ZJU Huzhou Institute, as well as plays the role of the Managing Editor for \textit{IET Cyber-Systems \& Robotics}. 
His research expertise is Flying Robotics, Control-theoretic Learning. Prof. Xu has published over 100 papers in international journals, including \textit{Science Robotics}, \textit{Nature Machine Intelligence}, etc.
\end{IEEEbiography}

\vspace{-0.7cm}
\begin{IEEEbiography}[{\includegraphics[width=1in,height=1.25in,clip,keepaspectratio]{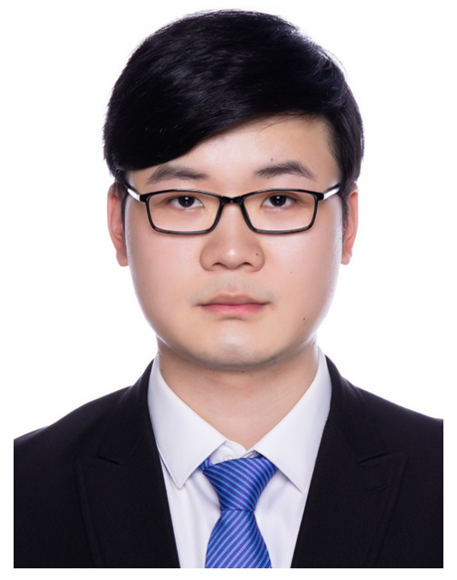}}]{Fei Gao}
received the Ph.D. degree in electronic and computer engineering from the Hong Kong University of Science and Technology, Hong Kong, in 2019.  
He is currently a tenured associate professor at the Department of Control Science and Engineering, Zhejiang University, where he leads the Flying Autonomous Robotics (FAR) group affiliated with the Field Autonomous System and Computing (FAST) Laboratory. 
His research interests include aerial robots, autonomous navigation, motion planning, optimization, and localization and mapping. 
\end{IEEEbiography}

\end{document}